\documentclass{article}

\usepackage{arxiv}

\usepackage[utf8]{inputenc}
\usepackage[T1]{fontenc}
\usepackage[hidelinks]{hyperref}
\usepackage{url}
\usepackage{booktabs}
\usepackage{amsfonts}
\usepackage{amssymb}
\usepackage{nicefrac}
\usepackage{microtype}
\usepackage{graphicx}
\usepackage[round]{natbib}
\usepackage{doi}
\usepackage[export]{adjustbox}
\usepackage{algorithm}
\usepackage{algpseudocode}
\usepackage{enumitem}
\usepackage{booktabs}
\usepackage{siunitx}
\usepackage[skip=2em]{caption}
\usepackage{textcomp}
\usepackage{listings}
\usepackage[colorinlistoftodos]{todonotes}
\usepackage[frozencache,cachedir=.]{minted}
\usepackage{mdframed}
\usepackage{tabularx}
\usepackage{makecell}
\usepackage{subfig}
\usepackage{booktabs, multirow}
\usepackage{soul}
\usepackage{changepage,threeparttable}
\usepackage{array}
\usepackage{wrapfig}
\usepackage{titlesec}

\titleformat{\paragraph}[hang]{\normalfont\normalsize\itshape}{\theparagraph}{1em}{}
\titlespacing{\paragraph}{0pt}{1.25ex plus .5ex minus .2ex}{0.5ex plus .2ex}

\newcommand*\rot{\rotatebox{90}}

\newcolumntype{P}[1]{>{\centering\arraybackslash}p{#1}}

\newminted[prompt]{django}{breaklines, breakindent=0pt, breaksymbolindentleft=0pt, breaksymbol=, fontsize=\footnotesize}
\surroundwithmdframed[linewidth=.5pt, topline=false,rightline=false, bottomline=false]{prompt}

\newminted[tinyprompt]{django}{breaklines, breakindent=0pt, breaksymbolindentleft=0pt, breaksymbol=, fontsize=\scriptsize}
\surroundwithmdframed[linewidth=.5pt, topline=false,rightline=false, bottomline=false]{tinyprompt}

\setlistdepth{9}

\setlist[itemize,1]{label=$\bullet$}
\setlist[itemize,2]{label=$\bullet$}
\setlist[itemize,3]{label=$\bullet$}
\setlist[itemize,4]{label=$\bullet$}
\setlist[itemize,5]{label=$\bullet$}
\setlist[itemize,6]{label=$\bullet$}
\setlist[itemize,7]{label=$\bullet$}
\setlist[itemize,8]{label=$\bullet$}
\setlist[itemize,9]{label=$\bullet$}

\renewlist{itemize}{itemize}{9}

\newenvironment{tabulars}[1]{
\renewcommand{\arraystretch}{1.3}%
\tabularx{\textwidth}{#1}}
{\endtabularx}

\definecolor{codegreen}{rgb}{0,0.6,0}
\definecolor{codegray}{rgb}{0.5,0.5,0.5}

\definecolor{backcolour}{RGB}{245,248,250}
\definecolor{emph}{RGB}{166,88,53}
\definecolor{nightblue}{RGB}{9,49,105}
\definecolor{keywords}{RGB}{207,33,46}
\definecolor{lightpurple}{RGB}{130,81,223}

\lstdefinestyle{gh}{
    backgroundcolor=\color{backcolour},   
    commentstyle=\color{codegreen},
    keywordstyle=\color{keywords},
    stringstyle=\color{nightblue},
    basicstyle=\ttfamily\footnotesize,
    breakatwhitespace=false,         
    breaklines=true,                 
    captionpos=b,                    
    keepspaces=true,                 
    showspaces=false,                
    showstringspaces=false,
    showtabs=false,                  
    tabsize=2,
    frame=shadowbox,
}

\title{Iterated Decomposition: Improving Science Q\&A \\ by Supervising Reasoning Processes}
\date{}

\author{ Justin Reppert\thanks{Correspondence to \texttt{justin@ought.org} and \texttt{andreas@ought.org}.}, Ben Rachbach, Charlie George, Luke Stebbing, \\
\textbf{ Jungwon Byun, Maggie Appleton, Andreas Stuhlmüller } \\ \\ Ought}

\renewcommand{\headeright}{}
\renewcommand{\undertitle}{}
\renewcommand{\shorttitle}{Iterated Decomposition: Improving Science Q\&A by Supervising Reasoning Processes}

\hypersetup{
pdftitle={Iterated Decomposition: Improving Science Q\&A by Supervising Reasoning Processes},
pdfsubject={cs.CL, cs.AI, cs.LG},
pdfauthor={Justin Reppert, Ben Rachbach, Charlie George, Luke Stebbing, Jungwon Byun, Maggie Appleton, Andreas Stuhlmüller},
pdfkeywords={Language models, question-answering, process supervision, alignment, transparency},
}

\begin{document}
\maketitle

\begin{abstract}
Language models (LMs) can perform complex reasoning either end-to-end, with hidden latent state, or compositionally, with transparent intermediate state. Composition offers benefits for interpretability and safety, but may need workflow support and infrastructure to remain competitive. We describe {\em iterated decomposition}, a human-in-the-loop workflow for developing and refining compositional LM programs. We improve the performance of compositions by zooming in on failing components and refining them through decomposition, additional context, chain of thought, etc. To support this workflow, we develop ICE, an open-source tool for visualizing the execution traces of LM programs. We apply iterated decomposition to three real-world tasks and improve the accuracy of LM programs over less compositional baselines on held-out test sets: describing the placebo used in a randomized controlled trial ($25\% \rightarrow 65\%$), evaluating participant adherence to a medical intervention ($53\% \rightarrow 70\%$), and answering NLP questions on the \textsc{Qasper} dataset ($38\% \rightarrow 69\%$). These applications serve as case studies for a workflow that, if automated, could keep ML systems interpretable and safe even as they scale to increasingly complex tasks.
\end{abstract}

\section{Introduction}

Language models are often trained using feedback on outcomes, leveraging reward models that imitate human evaluations provided as pairwise comparisons \citep{christiano2017, ziegler2020}. This works well for basic question-answering, summarization, simple code generation, and general short-form instruction-following \citep{stiennon2020, wu2021, ouyang2022}. For these tasks, good outputs can readily be distinguished from bad ones, especially with model-supported evaluation \citep{saunders2022}. And with rare exceptions like WebGPT \citep{nakano2021}, there is no difference between the model's output and the relevant outcomes, so evaluations relatively directly inform the model's behavior.

However, as model capabilities and task complexities scale up, outcome-based evaluation may run into alignment problems:

First, for some important applications the process used to generate the output matters as much as the output itself. Consider long-range forecasting, and policy decisions informed by such forecasts: The quality of a forecast or decision depends on the assumptions, evidence, and reasoning used to produce it. If feedback doesn't inform the process used to generate outputs, we may get results that look good (because they are optimized for favorable evaluation), and may even \emph{look} systematic, but are worse in exactly the ways that can't easily be measured \citep{stuhlmuller2022}.

Second, outcome-based feedback may create incentives for language models to deceive or manipulate their users by exploiting gaps or biases in the feedback signal \citep{amodei2016}. In extreme cases, this could lead to models behaving as intended during training, hiding their true intentions and capabilities, but defecting at deployment \citep{cotra2022, ngo2022}.

Process supervision is an alternative to outcome-based training that uses language models to execute human-understandable task decompositions, either by imitating human steps or by rewarding human-endorsed steps \citep{christiano2016, stuhlmuller2022, uesato2022}. This paradigm promises increased interpretability, trust, and alignment by reducing the reliance on black-box computation and enabling users to inspect and intervene in the model's reasoning process.

Right now, process supervision is ahead of outcome-based training: Language model capabilities are weak, so complex tasks require a composition of multiple calls. Indeed, engineered multi-step pipelines have been a staple of NLP for a long time. However, without scalable infrastructure to support it, we expect that outcome-based training will eventually crush process supervision performance-wise, leading to the alignment problems above.

This paper describes our experience applying process supervision to academic question-answering in the context of Elicit\footnote{\url{https://elicit.org}}, the AI research assistant developed by Ought. Our contributions are:

\begin{enumerate}
\item A review of the literature on process supervision, highlighting gaps in workflows and tooling that contribute to making real-world use cases rare
\item Iterated decomposition, a human-in-the-loop workflow for developing compositional language model programs
\item ICE, an open-source visualizer for language model execution traces
\item Case studies that use this workflow to improve performance over baselines on three real-world tasks: 
\begin{enumerate}
\item extracting placebo information from randomized controlled trials (RCTs), 
\item analyzing participant flow in RCTs, and 
\item answering questions about natural language processing papers.
\end{enumerate}
\end{enumerate}

As models advance and become more reliable at completing component tasks, we expect that process supervision will become more feasible, and that eventually the iterated decomposition process will be automated. By sharing our workflow and tooling now when even basic tasks are still challenging, we hope to accelerate a future where LM deployments are controllable and interpretable.

\section{Process Supervision}
\label{process_supervision}

\begin{figure}
    \centering
    \includegraphics[width=\textwidth]{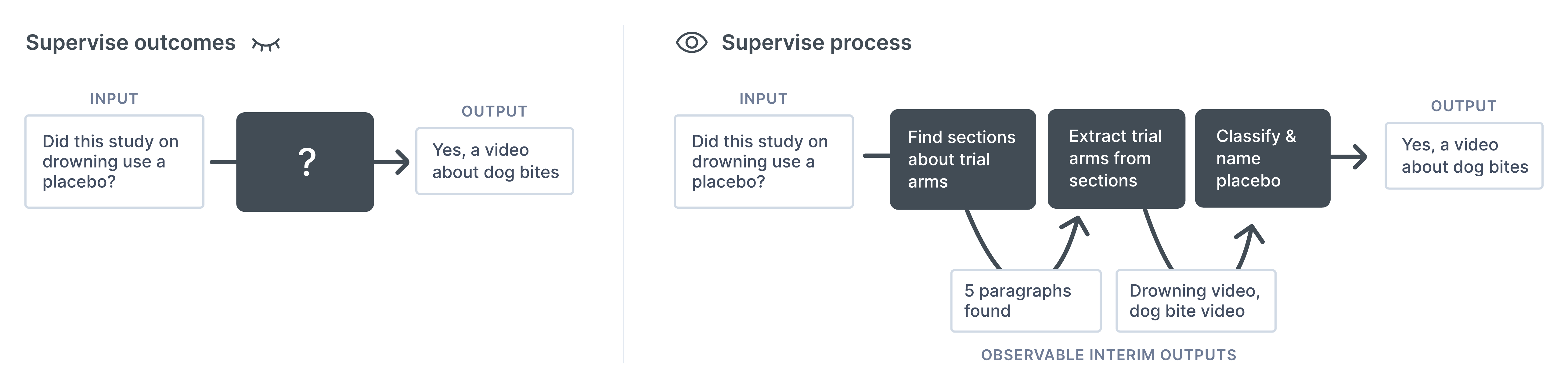}
    \caption{Process supervision breaks black-box language model computation into human-understandable reasoning steps without end-to-end optimization. This increases transparency and trust because it allows inspection of intermediate results and reduces risks from imperfect feedback signals and deceptive alignment.}
    \label{fig:process_outcome}
\end{figure}

Process supervision refers to approaches to LM training and deployment that rely on human-understandable intermediate steps. We review the literature and highlight gaps, then explain the \emph{iterated decomposition} workflow and ICE visualizer.

\begin{table}[!htp]\centering
\scriptsize
\setlength\extrarowheight{-2pt}
\begin{tabulars}{p{3cm}XP{.2cm}P{.2cm}P{.2cm}P{.2cm}P{.2cm}P{.2cm}}
Authors &Title &\rot{Decomposition} &\rot{Training} &\rot{Workflow} &\rot{Survey} &\rot{Tool} &\rot{Theory} \\\midrule
\citet{saha2022} &Summarization Programs: Interpretable Abstractive Summarization [..] &\checkmark &\checkmark &\checkmark & & & \\
\citet{wu2021} &Recursively Summarizing Books with Human Feedback &\checkmark &\checkmark &\checkmark & & & \\
\citet{yao2022} &ReAct: Synergizing Reasoning and Acting in Language Models &\checkmark &\checkmark & & & & \\
\citet{kumar2020} &NILE : Natural Language Inference with Faithful Natural Language Explanations &\checkmark &\checkmark & & & & \\
\citet{nakano2021} &WebGPT: Browser-assisted question-answering with human feedback &\checkmark &\checkmark & & & & \\
\citet{creswell2022} &Selection-Inference: Exploiting LLMs for Interpretable Logical Reasoning &\checkmark &\checkmark & & & & \\
\citet{jung2022} &Maieutic Prompting: Logically Consistent Reasoning with Recursive Explanations &\checkmark &\checkmark & & & & \\
\citet{sanyal2022} &FaiRR: Faithful and Robust Deductive Reasoning over Natural Language &\checkmark &\checkmark & & & & \\
\citet{creswell2022a} &Faithful Reasoning Using Large Language Models &\checkmark &\checkmark & & & & \\
\citet{bostrom2022} &Natural Language Deduction through Search over Statement Compositions &\checkmark &\checkmark & & & & \\
\citet{wang2022} &Locate Then Ask: Interpretable Stepwise Reasoning for Multi-hop Question Answering &\checkmark &\checkmark & & & & \\
\citet{fu2021} &Decomposing Complex Questions Makes Multi-Hop QA Easier and More Interpretable &\checkmark &\checkmark & & & & \\
\citet{dua2022} &Successive Prompting for Decomposing Complex Questions &\checkmark &\checkmark & & & & \\
\citet{guo2022} &Complex Reading Comprehension Through Question Decomposition &\checkmark &\checkmark & & & & \\
\citet{shridhar2022} &Distilling Multi-Step Reasoning Capabilities of LLMs into Smaller Models [..] &\checkmark &\checkmark & & & & \\
\citet{zhou2022a} &Learning to Decompose: Hypothetical Question Decomposition [..] &\checkmark &\checkmark & & & & \\
\citet{khattab2022} &Demonstrate-Search-Predict: Composing retrieval and language models [..] &\checkmark & &\checkmark & &\checkmark & \\
\citet{press2022} &Measuring and narrowing the compositionality gap in language models &\checkmark & & & & &\checkmark \\
\citet{khot2022} &Decomposed Prompting: A Modular Approach for Solving Complex Tasks &\checkmark & & & & & \\
\citet{ozturkler2022} &ThinkSum: Probabilistic reasoning over sets using large language models &\checkmark & & & & & \\
\citet{yang2022} &Re3: Generating Longer Stories With Recursive Reprompting and Revision &\checkmark & & & & & \\
\citet{gao2022} &PAL: Program-aided Language Models &\checkmark & & & & & \\
\citet{drozdov2022} &Compositional Semantic Parsing with Large Language Models &\checkmark & & & & & \\
\citet{trivedi2022} &Interleaving Retrieval with Chain-of-Thought Reasoning for [..] Multi-Step Questions &\checkmark & & & & & \\
\citet{kazemi2022} &LAMBADA: Backward Chaining for Automated Reasoning in Natural Language &\checkmark & & & & & \\
\citet{zelikman2022} &STaR: Bootstrapping Reasoning With Reasoning & &\checkmark &\checkmark & & & \\
\citet{xie2022} &Calibrating Trust of Multi-Hop Question Answering Systems with Decompositional Probes & &\checkmark &\checkmark & & & \\
\citet{uesato2022} &Solving math word problems with process- and outcome-based feedback & &\checkmark & & & & \\
\citet{dohan2022} &Language Model Cascades & & &\checkmark &\checkmark &\checkmark & \\
\citet{stuhlmuller2022a} &Factored Cognition Primer & & &\checkmark &\checkmark & & \\
\citet{wu2022a} &PromptChainer: Chaining Large Language Model Prompts through Visual Programming & & &\checkmark & &\checkmark & \\
\citet{wu2022} &AI Chains: Transparent and Controllable Human-AI Interaction by Chaining LLM Prompts & & &\checkmark & &\checkmark & \\
\citet{chase2022} &LangChain & & & &\checkmark &\checkmark & \\
\citet{polu2022} &Dust & & & & &\checkmark & \\
\citet{wies2022} &Sub-Task Decomposition Enables Learning in Sequence to Sequence Tasks & & & & & &\checkmark \\
\bottomrule
\end{tabulars}
  \caption{A sample of prior work on process supervision, categorized into work that primarily contributes (1) new task decompositions, (2) training and finetuning techniques, (3) workflows and tutorials, (4) surveys and frameworks for organizing prior work, (5) tools, and (6) theory. While work on decompositions and training techniques is rapidly growing, there is little investigation of workflows, tooling, and theory.}
  \label{tab:prior_work}
\end{table}

\subsection{Prior work on process supervision}

Over the past year, there have been significant advances in techniques for process supervision as well as frameworks, interfaces, and libraries for implementing it. At the same time, real-world use cases are still rare. We review prior work and highlight gaps in the literature that may be contributing to this. This is a rapidly growing field, so we are only able to review a sample of the work (Table \ref{tab:prior_work}).

\paragraph{Task decompositions} There is a quickly growing literature on how to compose multiple language model calls to improve performance or accomplish more difficult tasks:

\begin{itemize}
    \item \cite{creswell2022} and \cite{creswell2022a} generate reasoning steps by alternating between selection and inference. They show that their approach outperforms other prompting methods on multi-step logical deduction and scientific QA tasks, and generates interpretable reasoning traces.
    \item \citet{kazemi2022} apply backward-chaining to simple logic tasks, starting with a goal proposition and recursively decomposing it into sub-goals until the sub-goals can be proved or disproved.
    \item \cite{wu2021} and \cite{saha2022} apply recursive summarization to generate summaries of long texts, such as books or articles. They use LMs to summarize small sections of the text and then recursively summarize these summaries to produce a summary of the entire text. They show that recursive summarization improves the quality and coherence of the summaries, and enables human feedback and evaluation.
    \item \cite{yang2022} generate long stories by first creating a story plan, generating passages b yprompting a model with contextual information from the plan and the current story state, and then revising the passages by reranking and editing them.
    \item ReAct \citep{yao2022} interleaves generating chain-of-thought reasoning and actions pertaining to a task (e.g., search, lookup).
    \item WebGPT \citep{nakano2021} finetunes LMs to answer long-form questions using a text-based web-browsing environment.
    \item \citet{gao2022} offload computation to Python interpreters. 
    \item \citet{trivedi2022} and \citet{khattab2022} interleave chain-of-thought with knowledge retrieval steps.
    \item \citet{khot2022} study decomposition in general, letting the language model push subtasks to task-specific handlers.
    \item \citet{ozturkler2022} aggregate language model probabilities using mathematical combinators like sum and product.
    \item \citet{jung2022} recursively generate a tree of explanations for a statement, then determine the truth of the statement by treating the inference as a satisfiability problem over these explanations and their logical relations.
    \item Various works, including \citet{fu2021}, \citet{dua2022}, \citet{guo2022}, and \citet{zhou2022a}, explore decomposition of questions into subquestions, often under the name multi-hop question-answering.
\end{itemize}

\paragraph{Training and finetuning techniques}
The most relevant work is \cite{uesato2022} who directly compare process-based and outcome-based feedback for solving math word problems with LMs. They find that process-based feedback improves the correctness and interpretability of reasoning steps, but requires more label supervision than outcome-based feedback.

\citet{wu2021} finetune GPT-3 using behavioral cloning and reward modeling to do summarization recursively.
\citet{zhou2022a} train a decomposition model based on a parallel news corpus.
\citet{nakano2021} compare behavior cloning, RL and rejection sampling for training a web browsing model, and find that a combination of behavior cloning and rejection sampling against a reward model worked best.
\citet{zelikman2022} generate chains-of-thought, finetuning on the ones that lead to correct answers.

Most other work under ``training'' in Table \ref{tab:prior_work} trains small models from scratch for particular composition steps.

\paragraph{Theory and conceptual advances}
\citet{wies2022} show some benefits of process supervision: When concatenating intermediate supervision to the input and training a sequence-to-sequence model on this modified input, unlearnable composite problems can become learnable.

\cite{press2022} coin the name ``compositionality gap'' for the fraction of questions that the model answers incorrectly out of the questions for which the model answers all of the sub-questions correctly. They find this number to be around 40\%. They show that chain of thought can narrow the gap, and that generating and answering follow-up questions further narrows it.

\paragraph{Workflows, tutorials, and tools}
\cite{dohan2022} introduce language model cascades, which are probabilistic programs that compose LMs with random variables and control flow. They formalize several existing techniques, such as scratchpads, verifiers, STaR, selection-inference, and tool use, as instances of language model cascades. They provide an open-source probabilistic programming system.
The Factored Cognition Primer \citep{stuhlmuller2022a} is a tutorial that explains how to write compositional LM programs, including recursive question-answering, debate, search, and verification.
\citet{xie2022} show in human participant studies that letting users probe a language model with subquestions helps them calibrate when the model is correct.
\citet{wu2022a} and \citet{wu2022} describe a closed-source visual programming interfaces for making compositional language model programs.
Dust \citep{polu2022} is a web service and Rust library for designing and deploying LM apps.
LangChain \citep{chase2022} is a Python library that assists in the development of LM applications that involve chaining LMs with each other or with other experts.

\paragraph{Relation to this work} While there is a quickly growing literature on task decompositions, there is effectively no work on theory and only a few tools, tutorials, and workflows for building real-world process-based systems. We see this paper as demonstrating a real-world application (science Q\&A) as well as a description of the workflow (iterated decomposition) and tooling (ICE) used.

The prior work above varies in what exactly is meant by ``process'' and ``supervision''. We provide a taxonomy in Appendix \ref{sec:taxonomy}. Briefly, in this work, we focus on decompositions designed by a human developer, with occasional choices made by the language model. We balance pragmatic decompositions that improve task performance with decompositions that reflect ideal reasoning and facilitate better supervision. Our decompositions mostly improve performance by helping the model use long context more effectively, although some also apply multiple lines of reasoning to subtasks. When we talk about ``supervision``, we mean checking the outputs or behavior of individual steps in the composition. We supervise the process to improve the LM program, the supervisor is the human developer, and we always have access to the correct answer.

\subsection{Iterated decomposition}

\begin{figure}
    \centering
    \includegraphics[width=\textwidth]{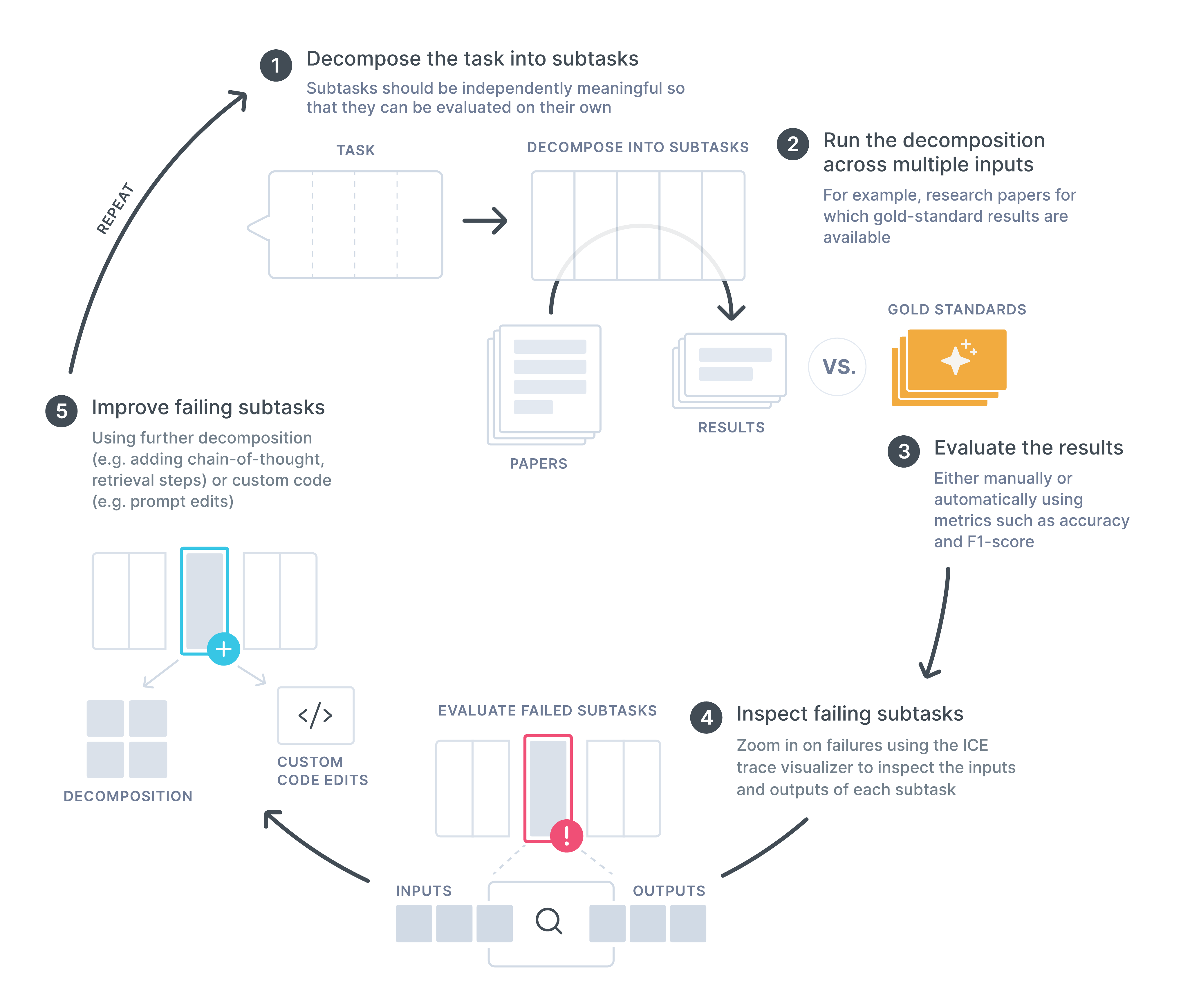}
    \caption{Iterated decomposition is a workflow for human-in-the-loop language model programming. We start with a trivial decomposition, evaluate it against gold standards, diagnose the source of failures using the ICE visualizer, refine the failing subtasks through further decomposition or other adjustments, and repeat.}
    \label{fig:workflow}
\end{figure}

We study process supervision via {\em iterated decomposition}, a human-in-the-loop workflow that incrementally improves a task decomposition through error diagnosis and amendment (illustrated in Figure \ref{fig:workflow}):

\begin{enumerate}
    \item Start with a minimal decomposition, breaking the task into subtasks that can be performed by a LM. For example, the task of extracting the placebo used in academic studies can be decomposed into first finding the most relevant section, then generating the placebo (if any) given that section.
    \item Apply the decomposition to multiple test inputs with gold standard answers. For example, this could be a dataset of academic papers from various domains, such as medicine, psychology, and economics, and their corresponding placebo descriptions, if applicable.
    \item Evaluate the results automatically (using LM) or manually, using metrics such as accuracy, F1-score, or other metrics. For example, the result of the placebo extraction task can be compared to a gold standard dataset of placebo descriptions from academic papers.
    \item Identify failures using the ICE trace visualizer or other tools to inspect the intermediate inputs and outputs of each subtask. For example, given that the generated placebo is incorrect, one can determine whether the wrong section was found or the right section was found but the wrong answer was extracted.
    \item Improve failing components using further decomposition (e.g., semantic decomposition into subtasks, chain-of-thought, retrieval), or by making custom edits (e.g. prompt tweaks, language model inference parameters). For example, the task can be split into first finding each experiment mentioned in the paper, then asking for each experiment whether it used a placebo or not.
\end{enumerate}

The process repeats steps 2-5 until it achieves good performance or exhausts the relevant resources (time, compute budget for experiments).

\subsection{Interactive Composition Explorer}

\begin{figure}
    \centering
    \includegraphics[width=\textwidth]{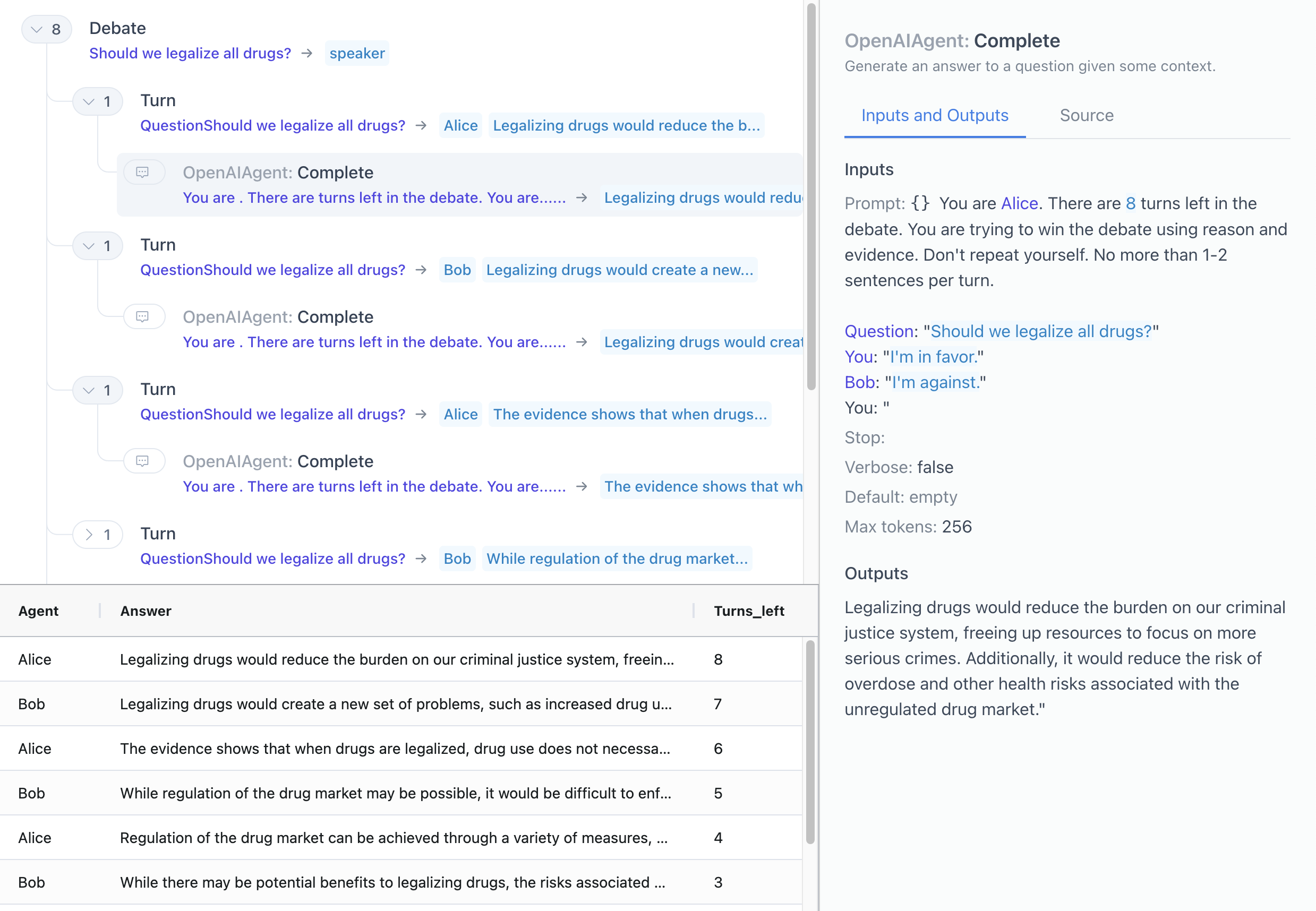}
    \caption{The Interactive Composition Explorer (ICE) is an open-source debugger for language model execution traces. It visualizes task decompositions and supports zooming in on failing subcomponents. The top left pane in the screenshot shows the trace of function calls in the original execution hierarchy. The bottom pane supports sorting and filtering of calls, with each row showing values that were recorded at execution time. The right pane shows function inputs, outputs, recorded intermediate values, and source code of the call selected in the main pane.}
    \label{fig:trace_example}
\end{figure}

To support the error identification step in iterated decomposition, we have developed the Interactive Composition Explorer (ICE), an open-source\footnote{\url{https://github.com/oughtinc/ice}} debugger and execution trace visualizer for language model programs.

ICE records and visualizes the execution of all async functions in a Python program, including all inputs (prompts, parameters) and outputs (language model responses, intermediate values), in a way that is suitable even for executions with hundreds to thousands of LM calls. ICE provides a decorator that can be applied to any async function that should be recorded, as well as utilities for recording all top-level async functions defined in a given module, any custom values of interest within a function, and the structure of each interpolated string. This data can then be visualized in a browser for interactive exploration, as shown in Figure \ref{fig:trace_example}.

ICE provides three views into this data:

\begin{enumerate}
  \item An expandable tree of function calls shown in order of execution
  \item A sortable, filterable table of function calls and their custom values
  \item A function call detail pane containing inputs, custom values, outputs, and source code
\end{enumerate}

The tree can be browsed to a particular function call which can then be inspected in the detail pane. Often, it is helpful to compare many calls to the same function, so ICE provides a dropdown menu of all recorded functions in the execution trace and their respective call counts. If a function is selected, all calls to it will be shown in the table and highlighted in the tree.

Since prompts are of particular interest in LM programs, the detail pane includes special support for rendering interpolated strings (f-strings). Each interpolated value is shown in an alternating color, and the interpolating source code is shown in a tooltip.

These functions all serve the purpose of understanding, analyzing, and debugging task decompositions, going from the high-level structure of the call tree to seeing all functions of a particular type in the table, to seeing what exactly the prompt was when a particular instance of that type generated an unexpected result.

\section{Real-World Context of Case Studies}

\begin{figure}
    \centering
    \includegraphics[width=450pt]{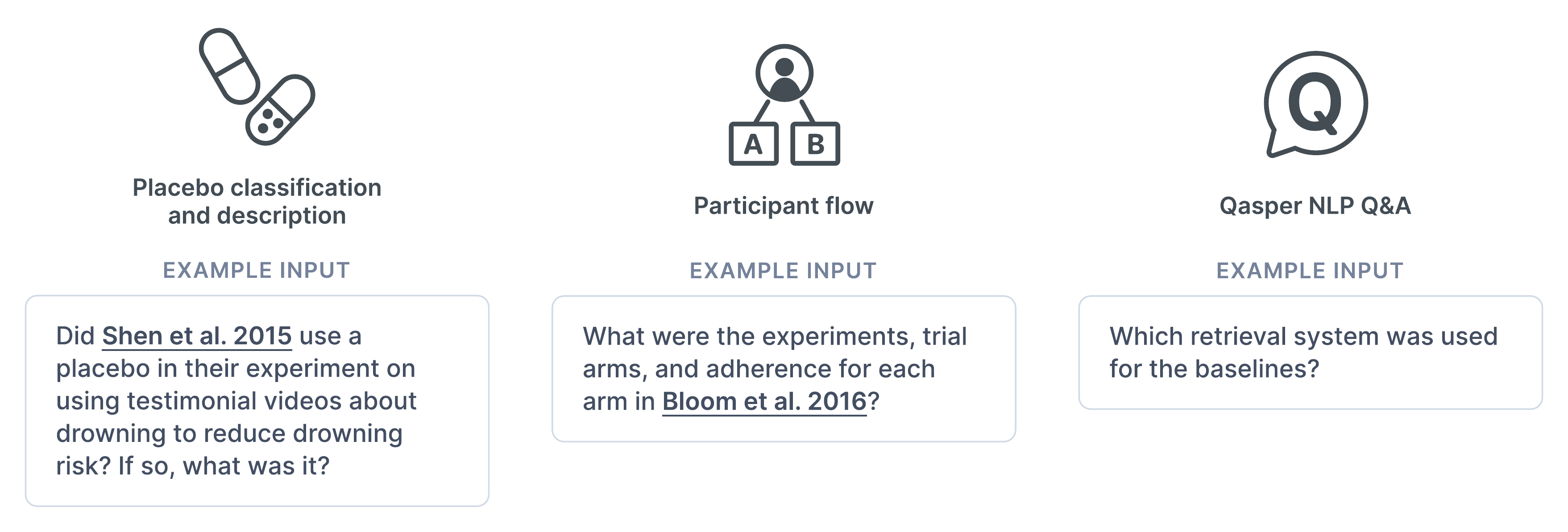}
    \caption{We apply iterated decomposition to three real-world case studies: Extracting placebos from academic papers about randomized controlled trials, extracting participant flow, and NLP question-answering. The tasks are part of our work on automated literature review for the AI Research Assistant Elicit.}
    \label{fig:all_tasks}
\end{figure}

We used iterated decomposition and ICE to improve performance over simple baselines on three real-world case studies we encountered in our work on Elicit (Figure \ref{fig:all_tasks}).

A primary use case for Elicit is to find out what the academic literature knows about a research question. Many of Elicit's users work in biomedicine, psychology, experimental economics, and other fields where randomized controlled trials (RCTs) are an important source of evidence. When these users do literature review or metanalyses, they often roughly follow this process:

\begin{enumerate}
\item What are the RCTs that are potentially most relevant to answering my research question?
\item For each RCT:
  \begin{enumerate}
  \item Does this study actually address my research question?
  \item What is the risk of bias? Can I take the findings at face value, or do I need to discount it or ignore it?
  \end{enumerate}
\item What does the aggregate of RCTs, weighted by evidence quality, say about my research question?
\end{enumerate}

The \textbf{Placebo Classification $\&$ Description} task is designed to help researchers evaluate risk of bias. Information about the placebo helps researchers assess the risk of bias in the study and decide how much to trust the results – comparing an intervention to a placebo control is usually stronger evidence than comparing to a no-treatment control, especially if the placebo successfully blinded the participants.

The first step of the \textbf{Participant Flow in RCTs} task is to identify the trials and trial arms within each study. This helps the user understand what the researchers did in the study, to determine whether it addresses their research question. The second step, evaluating participant adherence to an intervention, helps the researcher assess risk of bias and contextualize the study's findings: Was a small effect driven by poor adherence to the treatment?

These tasks represent a much broader class of tasks in Elicit: answering questions about papers. Elicit currently supports about 20 pre-specified questions, but also allows users to enter their own questions. Ultimately, we want to find highly generalizable strategies to improve performance. The \textbf{\textsc{Qasper} NLP Q$\&$A} task tests this by generalizing the participant flow decomposition to a different domain.

\section{Case Study: Placebo Classification \& Description}
\label{sec:case-study-placebo}

In this case study, we focus on domain-specific decompositions. Through iteration on the classification and description steps, accuracy of the generated placebo description improved from $25\%$ to $65\%$ on a held-out test set.

\subsection{Setup}

Given the full text of an RCT, the task is to answer ``Did this RCT use a placebo?'' (classification) and ``If so, what was it?'' (description). For both classification and description we find strong agreement among human raters.

Appendix \ref{sec:appendix_placebo_task} has a detailed description of the task. Table \ref{tab:placebo_examples} in the Appendix shows examples of the most relevant quotations from two papers with corresponding classifications and descriptions. 

\begin{figure}
  \centering
  \includegraphics[width=\textwidth]{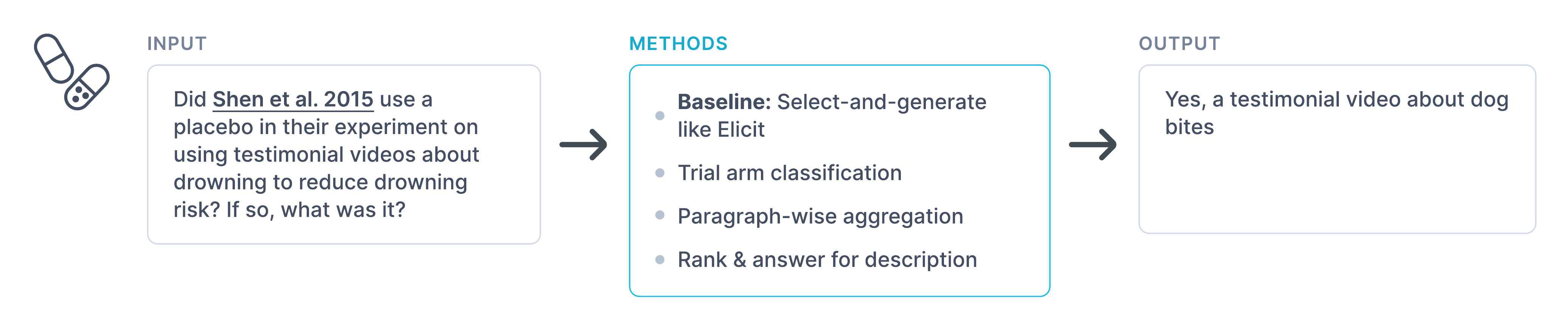}
  \caption{In the placebo case study, the goal is to determine if a randomized controlled trial used a placebo, and if so, what it was. We start with the select-and-generate baseline used in Elicit and improve it by ensembling classification based on trial arms and paragraphs.}
  \label{fig:placebo_task}
\end{figure}

\subsection{Iterations}

\subsubsection{Baseline: Select-and-generate like Elicit}
\label{sec:elicit-baseline-explanation}

The baseline is the ``select-then-generate'' algorithm currently deployed for paper question-answering in Elicit. Elicit uses monoT5\footnote{\url{https://huggingface.co/castorini/monot5-base-msmarco-10k}} to rank paragraphs from the paper against the question \citep{lin2021}. The top-ranking paragraph is fed to \texttt{text-davinci-002} with the following prompt:

\begin{prompt}
Answer the question "{{question}}" based on the excerpt from a research paper. 
Try to answer, but say "The answer to the question is not mentioned in the excerpt" if you don't know how to answer. 
Include everything that the paper excerpt has to say about the answer. Make sure everything you say is supported by the excerpt.
The excerpt may cite other papers; answer about the paper you're reading the excerpt from, not the papers that it cites. 
Answer in one phrase or sentence:
Paper title: {{title}}
Paper excerpt: {{paragraph}}
Question: {{question}}
Answer:
\end{prompt}

On a held-out test set, this approach has 71\% accuracy for classification, 25\% for description (Table \ref{tab:placebo_results}).

\subsubsection{Improving classification}

\paragraph{Diagnosis} For each of the 14 trials that the baseline classified incorrectly, there was in fact a placebo. In all cases, the selection step failed to find the part of the paper that most clearly stated that there was a placebo. For 10/14 trials, selection found no evidence that there was a placebo. For the remaining 4/14 trials, selection found some mention of the placebo but not a full description, then generation failed to conclude that there was a placebo (see Appendix \ref{sec:appendix_elicit_selection_failures} for examples).

To improve classification, we needed a way to more robustly find evidence that there was a placebo without introducing false positives.

\paragraph{Solution} We created a decomposition that mirrors our own reasoning process for determining placebo classification and description. When we read the papers in our validation set, we noticed two things: First, almost all papers clearly describe their trial arms. Then, if there was a placebo in the study, one of the arms will usually be clearly identified as the placebo arm. So, we identify and describe each trial arm and classify whether any of the arms is a placebo. Second, often many paragraphs in the paper provide evidence about whether the paper used a placebo, and sometimes this evidence is contradictory. So, we could check what each paragraph tells us about whether the paper used a placebo, then aggregate those answers.

In outline, the decomposition looks like this:

\begin{enumerate}
\item Classify based on trial arms
\begin{enumerate}
\item What were the trial arms?
  \begin{enumerate}
    \item Rank paragraphs by relevance to this question through pairwise comparisons
    \item Answer based on the most relevant paragraphs
  \end{enumerate}
\item Describe each trial arm: Rank and Answer
\item Do any of the arms look like placebos?
\item If so, could participants tell which arm they were in? (If so, there's not really a placebo.)
\end{enumerate}
\item Classify based on each paragraph
\begin{enumerate}
\item Was there a placebo, according to each paragraph?
\item Aggregate answers from paragraphs
\end{enumerate}
\item Ensemble the classification based on trial arms and paragraphs
\end{enumerate}

Appendix \ref{sec:appendix_placebo_decomposition} shows the full decomposition.

\paragraph{Results} This approach substantially improves on the baseline (71\% correct $\rightarrow$ 98\%; p=0.0004), achieving near-perfect performance (Table \ref{tab:placebo_results}).

Reviewing the execution traces in ICE showed that the improved performance is almost entirely due to answering based on each paragraph, not due to answering based on trial arms. Whenever the arms-based classification tentatively classified one of the trial arms as a placebo, it also found that participants might be able to tell which arm they were in, and so it ultimately said that it was unclear whether there was a placebo.

\subsubsection{Improving description}

\paragraph{Diagnosis} 14 of the 15 trials where the baseline failed to describe the placebo correctly are ones that it failed to classify correctly. The baseline does not even attempt to describe the placebo if the classification is ``no placebo'', so it fails all of these by construction. However, in each of these cases selection failed to find the excerpts needed for description, so any attempt at description would have failed.

\paragraph{Solution} We noticed that the information required to describe the placebo is often dispersed throughout the paper. So, we rank the most relevant paragraphs in the paper and then use them to generate an answer. We re-use the same Rank and Answer technique that we used for classification. This generalizable subtask decomposition works well for both purposes.

\paragraph{Results} This approach results in a big improvement on the Elicit baseline (25\% correct \( \rightarrow \) 65\%; \(p=0.025\)---see Table \ref{tab:placebo_results}).

\begin{table}
  \centering
    \begin{tabulars}{XSS}
      \toprule
      \textbf{Method} & {\textbf{Classification (n=48)}} & {\textbf{Description (n=20)}} \\ 
      \midrule
      Elicit select-then-generate baseline & \text{71\%} & \text{25\%} \\ 
      Select-then-generate with “not mentioned" perplexity selection & \text{65\%} & \text{10\%} \\
      Stuff paper in prompt & \text{69\%} & \text{30\%} \\ 
      Final decomposition & \bfseries{98\%} & \bfseries{65}\% \\      
      Keyword decision tree & \text{94\%} & \text{10\%} \\       
      \bottomrule
    \end{tabulars}
  \caption{Through iteration on the task decomposition, accuracy of the placebo classification improved from $71\%$ to $98\%$ and description accuracy improved from $25\%$ to $65\%$, both on a held-out test set.}
  \label{tab:placebo_results}
\end{table}

\subsubsection{A keyword baseline}

The strong paragraph-based classification results raise the question whether can we encode our understanding of how to figure out whether a trial used a placebo in a much simpler algorithm.

We created a simple regex keyword-matching algorithm:

\begin{enumerate}
\item Classify as no placebo if the paper contains words like "open-label"
\item Then classify as placebo is the paper contains words like "placebo", and take the first matching sentence as a description
\item Then classify as no placebo if none of these words are present
\end{enumerate}

This algorithm does about as well as our task decomposition on classification. However, a regex keyword approach doesn't generalize to harder and more ambiguous tasks---for example, adherence is discussed using a variety of wordings, so keywords don't help as much to select relevant passages (Appendix \ref{sec:adherence_examples}).

\section{Case Study: Participant Flow in Randomized Controlled Trials}
\label{sec:case-study-participant-flow}

In this case study, we mostly focused on simple generalizable decompositions for long-form question-answering. Through iteration on the selection and generation steps, accuracy on all subtasks improved substantially on a held-out test set: Extracting experiments improved from $40\%$ to $70\%$, trial arms from $55\%$ to $86\%$, and adherence from $53\%$ to $70\%$.

\subsection{Setup}

\begin{figure}[h]
    \centering
    \includegraphics[width=\textwidth]{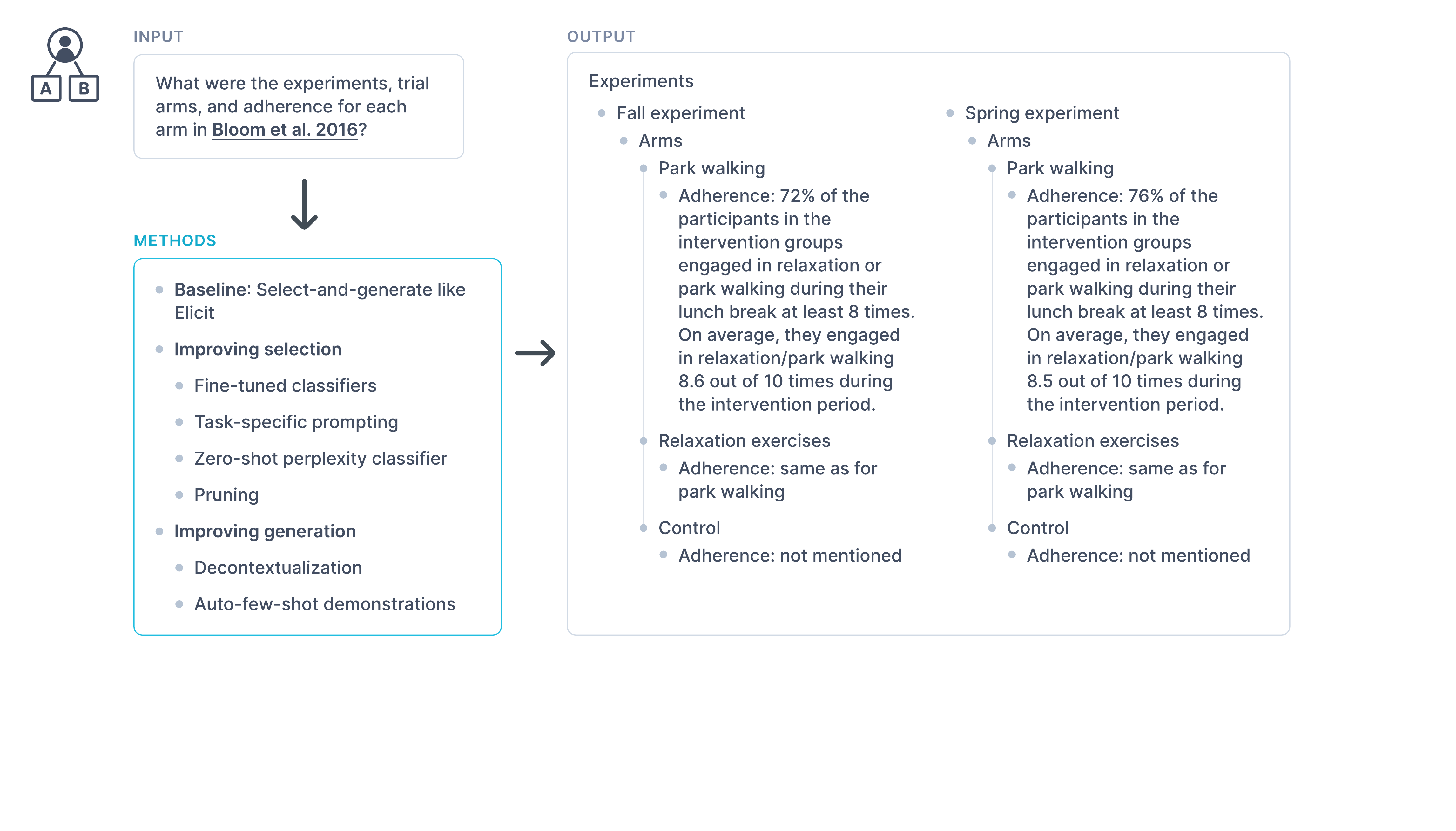}
    \caption{In the participant flow case study, the goal is to analyze a randomized controlled trial and extract the experiments, the arms for each experiment, and the adherence rate for each arm. Starting again from the Elicit select-and-generate baseline, we implemented domain-agnostic improvements to both selection and generation.}
    \label{fig:task_CONSORT}
\end{figure}

Randomized controlled trials often include a standardized diagram that helps the reader understand what happened in a given trial: How did participants journey through the study, from enrollment to final analysis? These diagrams are called \emph{CONSORT} diagrams (see Appendix \ref{sec:CONSORT} for an example). In our experience, they appear in about half of recent RCTs are very often incomplete. If we could generate complete CONSORT diagrams for all RCTs, we could provide valuable information to readers.

In this case study we consider a limited version of this task: We name the trials in the paper, the arms in each trial, and describe the participant adherence rate for each arm. Figure \ref{fig:task_CONSORT} shows an example.

\subsection{Evaluation}

Experiments and arms are relatively easy to evaluate---we can check whether each experiment/arm from the gold standard is represented and whether there are any additional ones that should not be there.

Adherence tends to require a narrative answer, and to be more subjective. This means that it may benefit more from a nuanced decomposition, and approaches that work well for adherence may generalize better to other fuzzy tasks.

Each arm has an adherence answer, so there are a total of 135 adherence answers in our test set. Often, no information about adherence is available in the text of the paper---information about adherence is only available for 56/135 arms (41$\%$) in our test set. So the adherence task is substantially a classification task (adherence mentioned vs.\ not mentioned).

\subsection{Iterations}

\subsubsection{Baseline: Select-and-generate like Elicit}

The baseline is the same ``select-then-generate'' approach used for general paper question-answering in Elicit as described above in Section \ref{sec:elicit-baseline-explanation}. This is a reasonably strong baseline, scoring \(53\%\) on the adherence subtask. 

\paragraph{Diagnosis}

For adherence, 80\% of errors were false negatives, i.e.\ saying that adherence was not mentioned when it in fact was (see Table \ref{tab:adherence_error_analysis}). Further, all (51/51) of these false negatives resulted from errors at the selection stage---the answer really was not mentioned in the top-1 paragraph from the monoT5 ranker. By using an oracle for selection with the same generation approach, accuracy on the adherence task rose to \(77\%\). So, the baseline fails primarily by failing to make good use of the long context of the paper. For this reason, we started by iterating on selection.

\subsubsection{Improving selection}

\paragraph{Finetuned classifiers}

Because our research is aimed at generalizable approaches, we restricted any finetuning to approaches that were either very sample-efficient (so little labeling effort would be needed to adapt the approach to new tasks) or very general. We evaluated small dataset and machine-in-the-loop labeling approaches, favoring approaches using at most a few hours of subject matter expert labeling effort.

We adapted the approach from \cite{liu2022} to fine tune classifiers for text classification subtasks. This approach combines low-rank adaptation (scaling selected activations by learned vectors) with additional regularization terms in the loss and hyperparameters the authors found to generalize well. This combination enables (a) a very small set of finetuned weights, enabling multiple finetuned tasks to share a single model backbone at inference time, and (b) true low-data finetuning, since the approach prescribes hyperparameters, thus obviating the need for separate training and validation sets.

\cite{liu2022} suggest that their approach works well with as few as 20 examples. bootstrapped to a larger number of positive examples and more than 20 "hard negative" examples using weak models: For our tasks, most paragraphs are not relevant to answering a given question, and it is time-consuming and expensive to collect even a moderate amount of positive examples by hand. We ensembled multiple weak classifiers such as monoT5, BART-based classifiers, and T0-3B to identify positive examples from thousands of papers, then had experts moderate the examples identified by all the weak classifiers as answering the question, leaving in any negatives from this approach as ``hard negatives''. Intuitively, we would expect that this would provide an outsized number of ``easy positives'' (from the ensemble consensus) and ``hard negatives'' (false positives from the ensemble consensus, as corrected by experts). Although we would expect this approach to have a difficult time identifying ``hard positives'', in practice it created a dataset that worked very well in concert with T-few finetuning, bringing performance on a holdout set on the selection subtask for adherence from \(72\%\) before finetuning to \(94\%\) accuracy after.

Using T0-3B and T0pp (11B parameters) as base models, we finetuned using the T-Few approach. We found that the models generalized better when we left zero-shot prompts in our finetuning data, which can be interpreted as a form of regularization (in this example, the decoder answer choices are \texttt{yes} and \texttt{no}):

\begin{prompt}
Context: {{paragraph}}

Section: {{section}}

Answer yes if the following sentence is about how many participants in the study complied with the study's protocol, had to drop out, or withdrew; answer no if it is about something else, such as the study's design, sampling strategy, or results.

Sentence: {{sentence}}

Answer:
\end{prompt}

Many failures to get the right answer using finetuned classifiers at the selection stage were rooted in false negatives from the classifier (for 67\% of failures in the development set with \(n=9\), the classifier missed the gold standard excerpts), and the generation stage was somewhat robust to ``distractors'' caused by lower precision. We therefore explored variations on this technique to favor recall. For instance, we explored classifying \emph{sentences} (while including the paragraph as context), then, at inference time, classifying a paragraph as positive if any of its sentences were classified as positive. We also tried finetuning with multiple zero-shot prompt templates, then, at inference time, classifying a paragraph as positive if it was classified as positive via any of the prompt templates.

On the adherence subtask, the finetuned classifier significantly outperformed the base monoT5 classifier (paragraph classification \(F_1=0.10\) for Elicit top-1 approach versus \(F_1=0.77\) via T-Few finetuning of T0-3B).

Overall, while this approach led to significant performance improvement, it also had downsides: It requires data and training, and is less interpretable than some of the classification approaches we explored later. In the future, we would consider data-efficient finetuning of classifiers where (a) the task was hard to specify but data was easy to collect or (b) where the inference-time computational advantage of a smaller finetuned classifier was needed.

\paragraph{Task-specific prompting}

An alternative to finetuning is crafting prompts for specific subtasks. This reduces the data collection burden of finetuning, requiring just a few examples, at the cost of involving a different sort of expertise in the craft of writing effective prompts. 

With \texttt{text-davinci-002}, we found that we could get considerable performance gains on specific subtasks by prompt engineering. Naive zero-shot or few-shot instructions performed performed worse than ``simulation'' approaches where the prompt emulated a genre of text that had certain desirable properties, such as containing correct mathematical reasoning. Appendix \ref{sec:custom-adherence-prompt-textbook} shows an example of a prompt that frames a task as excerpts from a (nonexistent) textbook on evaluating scientific papers. Appendix \ref{sec:custom-adherence-prompt-stats-exchange} shows a prompt that simulates a StatsExchange post.

Because we wanted to focus on general-purpose decompositions in this case study, and because we expect simulation-based approaches to be less necessary with future instruction-tuned models, we did not evaluate this line of approaches systematically.

\paragraph{Zero-shot perplexity classifier}
\label{sec:zero_shot_perplexity}

Ideally, we could improve selection performance over the baseline monoT5 classifier with an approach that requires neither substantial data collection nor task-specific prompt engineering.

Inspired by the relative strength of the ``generation'' stage of the baseline approach, we implemented a zero-shot perplexity-based cross-encoder using \texttt{text-davinci-002} that performed remarkably well across many questions, outperforming the monoT5 classifier and competitive with the T-Few low-rank finetuning.

In this paragraph classification approach, we used the following prompt with \texttt{text-davinci-002}, echoing the log probs back from the OpenAI API in order to measure perplexity:

\begin{prompt}
Answer the question "{{question}}" based on the excerpt from a research paper. Try to answer, but say "The answer to the question is not mentioned in the excerpt" if you don't know how to answer. Include everything that the paper excerpt has to say about the answer. Make sure everything you say is supported by the excerpt. The excerpt may cite other papers; answer about the paper you're reading the excerpt from, not the papers that it cites. Answer in one phrase or sentence:
Paper excerpt: {{paragraph}}
Question: {{question}}
Answer: The answer to the question is not mentioned in the excerpt
\end{prompt}

We then scored paragraphs by the inverse perplexity of the tokens $X: \{(x_{i=1}, \ldots, x_t)\}$ in the sequence ``The answer to the question is not mentioned in the excerpt'':

\begin{equation}
    \exp\left(\frac{1}{t}\sum_{i=1}^{t}\log p_\theta(x_i|x_{< i})\right)
\end{equation}

We studied different uses of these scores. In the simplest approach using this perplexity-based method, we also generated the answer using this prompt, based on the paragraph with the highest perplexity for the ``not mentioned'' completion. This closely mimics the structure of the question-answering method used in Elicit, with the \texttt{text-davinci-002} perplexity-based ranker replacing the monoT5 ranker.

The \texttt{text-davinci-002} perplexity approach seemed more robust to subtle but important semantic differences than the monoT5 ranker. For example, on the adherence task, the monoT5 top-1 paragraph was often \emph{some} paragraph discussing adherence but not with respect to the specific trial arm in question. In many cases, the \texttt{text-davinci-002} perplexity approach correctly scored these ``about adherence but not the right trial arm'' paragraphs below results paragraphs about the correct trial arm that did not discuss adherence rates as explicitly but alluded to information about adherence for that arm.

Further, we found that it was generally better to include multiple paragraphs at the answering stage, in the order of the original paper. We adapted the perplexity top-1 approach into a classification approach by choosing a classification threshold empirically, slightly favoring recall over precision; the perplexity threshold adapted reasonably well across different questions, so we used the same threshold across all tasks. As expected, including multiple paragraphs above a threshold helped compensate for remaining weaknesses in the ranker, and, in some cases, multiple paragraphs were \emph{required} to generate a complete answer. Including multiple paragraphs instead of one paragraph mostly addressed cases where the top-1 approach incorrectly answered ``not mentioned''.

This zero-shot classification approach was remarkably strong for the ``select'' stage of our ``select-then-generate'' approaches, improving paragraph classification recall over the top-1 monoT5 Elicit baseline from \(0.09\) to \(0.76\) on the adherence task. It is zero-shot and requires no additional data (with the exception of needing data to set a classification threshold once across all tasks). The approach also offers a built-in interpretability benefit, in that we can look at the most likely completion that is not ``The answer to the question is not mentioned in the excerpt'' to understand what the model found relevant in each paragraph.

This approach also offers both classification and ranking via a natural interpretation and approach, where the ranks are reasonably well-calibrated; this means that, for instance, if we find that we have too many paragraphs to fit into the model's context window at the answering stage, we can with reasonable confidence simply remove lower-ranked paragraphs, without additional work.

\paragraph{Pruning}

Many implementations of the ``select'' stage both look at each paragraph independently and are tuned to slightly favor recall over precision. Conceptually, this may lead to a situation where we select redundant information or where, once we see all selected candidates together, it becomes clearer which ones contain the answer. Therefore, we tested a ``pruning'' step that looked at all paragraphs selected by the previous selection stage, in paper order, then reselected the paragraphs needed to answer the question.

Despite testing chain-of-thought and few-shot variants of this stage, our results here were only moderately promising. We were unable to substantially improve bottom line accuracy by adding this stage, and we were unable to improve paragraph classification precision without substantially hurting recall. For the experiments task, pruning improved precision from 0.14 to 0.29 while reducing recall from 0.77 to 0.54, increasing paragraph classification $F1$ from 0.24 to 0.38. This led to an improvement in final answer accuracy from 0.55 to 0.6. Further iteration on the implementation of this stage may yield better results.

\subsubsection{Improving generation}

Improving the ``select'' stage of the ``select-then-generate'' approach boosted performance on the participant flow tasks. However, the ICE traces revealed that the ``generate'' stage, which answers the question given the most relevant excerpts, also made interesting errors: Even with an oracle for selection, the baseline generation step only achieved \(77\%\) accuracy on the adherence task.

\paragraph{Error analysis} We identified some common errors in the generation stage when the selection stage had high recall.

Some errors came from lack of context in the selected paragraphs. They answered the question, but without the surrounding text, the answer was unclear. This happened often on the arms task: the selected paragraphs mentioned the trial arms, but not the corresponding trial or the experimental methodology, and so it was difficult to correctly describe the trial arms.

More often, the desired answer style and detail were unclear, especially for the adherence task. For example, some papers said ``adherence was generally good'' and gave numerical details, but the generation stage only included the former, not realizing the latter were also important.

Below we discuss two approaches we tried to reduce these errors.

\begin{table}
  \centering
    \begin{tabulars}{X p{1.9cm} p{1.2cm} p{1.4cm}}
      \toprule
      \textbf{Method} & {\textbf{Experiments (n=50)}} & {\textbf{Arms (n=50)}} & {\textbf{Adherence (n=135)}} \\ 
      \midrule
      Elicit select-then-generate baseline & \text{40\%} & \text{55\%} & \text{53\%} \\ 
      + “not mentioned" perplexity selection & \text{55\%} & \text{78\%} & \text{61\%} \\ 
      + “not mentioned" perplexity selection + prune & \text{60\%} & \text{70\%} & \text{54\%} \\ 
      + “not mentioned" perplexity selection + auto-few-shot selection & \text{65\%} & \text{72\%} & \text{55\%} \\ 
      + “not mentioned" perplexity selection + auto-few-shot selection + prune & \text{55\%} & \text{56\%} & \text{58\%} \\ 
      + “not mentioned" perplexity selection + auto-few-shot generation & \bfseries{70\%} & \bfseries{86\%} & \bfseries{70\%} \\ 
      + “not mentioned" perplexity selection + decontextualize & \text{65\%} & \text{75\%} & \text{54\%} \\      
      \bottomrule
    \end{tabulars}
  \caption{Through iteration on the participant flow decomposition, accuracy at extracting experiments improved from $40\%$ to $70\%$, trial arms from $55\%$ to $86\%$, and adherence from $53\%$ to $70\%$, all on a held-out test set.}
  \label{tab:CONSORT_flow_results}
\end{table}

\paragraph{Decontextualization} The select-then-generate pipeline may struggle to fully answer questions that require resolving coreferences or understanding context from surrounding text. To address this, we experimented with a decontextualization stage, inspired by \cite{choi2021}, that operates on the paper and adds context from previous paragraphs in square brackets. The goal is to make paragraphs more self-contained and easier to use for generation. We used an autoregressive model with a fixed few-shot prompt (see Appendix \ref{sec:decontext-prompt-appendix}) based on \citet{eisenstein2022a} to perform decontextualization. We did not systematically evaluate the accuracy or relevance of the added context, but we spot-checked some examples using ICE.

Decontextualization improved generation performance on some tasks, but not all. It helped the model identify the experiments and arms of the trials, as the added context clarified the methodology and the relevance of the selection. However, it did not help much with the adherence task, as most adherence information was already self-contained and the added context introduced more noise and confusion. The performance using decontextualized paragraphs as inputs to the generation stage was 0.65 for experiments and 0.75 for arms, compared to 0.40 and 0.55 for the baselines, respectively. For adherence, it was 0.54, barely above the baseline of 0.53.

\paragraph{Auto-generated ``demonstrations''} We studied the use of demonstrations to address a failure mode of the generation stage: producing answers with insufficient detail or inappropriate style. Demonstrations are examples of question-answering tasks with the same shape: (a) a question, (b) a text, (c) excerpts from that text that support an answer, and (d) an answer. Unlike other methods that encode expert knowledge, such as the placebo decomposition or the placebo keyword decision tree algorithm, demonstrations do not require any algorithm-specific or language-model-specific work from the expert. The expert simply gives examples of applying their expertise.

We generated few-shot prompts from expert demonstrations (gold standards) by prepending the existing perplexity selection prompt with multiple filled examples. We either used only positive demonstrations, which included a supporting paragraph and an expert's handwritten answer based on that paragraph, or mixed positive and negative demonstrations, which assumed that unmarked paragraphs did not answer the question. This assumption was often false, so we found it better to use only positive examples in the few-shot demonstration prompts.

Across all tasks, the few-shot prompts derived from demonstrations improved the accuracy of generation over the zero-shot approach, even with randomly sampled examples from the development set. The improvement was larger for harder tasks or tasks that benefited from implicit background knowledge. The demonstrations often helped by showing the expected types of answers for the task, such as including numerical detail where available. For example, for the adherence task, the zero-shot approach produced an incorrect answer like ``The paper says that both groups had good compliance with all products in each nursing home'', while the few-shot approach via demonstrations produced a correct answer like ``Compliance with oral supplementation was good, and daily intake averaged about 400 kcal''. The few-shot approach via demonstrations increased the final answer accuracy from $0.53$ to $0.70$ overall for this task.

\subsubsection{Combined improvements} The zero-shot perplexity classifier and few-shot answering via ``demonstrations'' add up to substantial improvements over the Elicit baseline (e.g. 53\% correct $\rightarrow$ 70\% for adherence, p=0.006\footnote{Fisher's exact test, considering each adherence answer to be independent. In reality the answers are clustered by trial and paper, so the true p-value is somewhat higher.}; see Table \ref{tab:CONSORT_flow_results}). These techniques vastly reduce the number of false negatives for the adherence task (error analysis in Appendix \ref{sec:participant-flow-error-analysis}).

\section{Case Study: \textsc{Qasper} NLP question answering}
\label{sec:case-study-qasper}

\begin{figure}
    \centering
    \includegraphics[width=\textwidth]{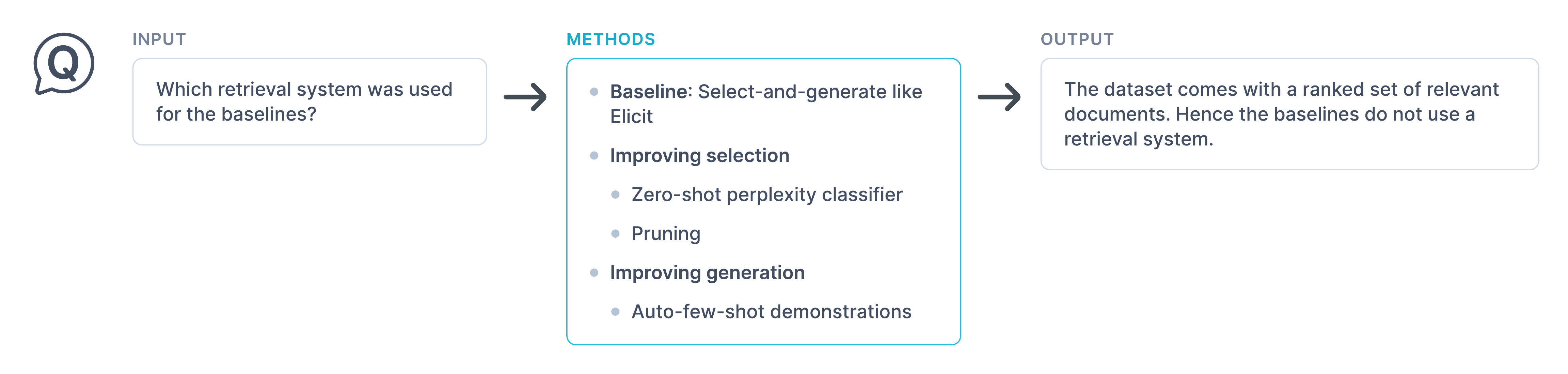}
    \caption{In the \textsc{Qasper} case study, the goal is to answer questions about NLP papers. \textsc{Qasper} is a dataset of 5049 information-seeking questions asked by regular readers of NLP papers and answered by separate NLP experts. Apart from finetuning we explored the same methods as for participant flow to learn how they generalize.}
    \label{fig:task_qasper}
\end{figure}

We primarily iterated on decompositions using the tasks above. However, we also wanted to explore how well these task decompositions generalized to other domains and questions, so we tested them on a subset of the \textsc{Qasper} dataset, a question-answering dataset of 5,049 questions and answers written by NLP practitioners, covering 1,585 NLP papers \citep{dasigi2021}. By using the final decomposition for participant flow, accuracy at answering NLP questions on the \textsc{Qasper} dataset improved from $38\%$ to $69\%$.

\subsection{Setup}

Answers to \textsc{Qasper} questions are in one of three formats: yes/no, excerpts from the text, or freeform answers. In the original \textsc{Qasper} paper and benchmark, the authors measured \(F_1\) scores of generated answers versus the expert answers. Because we care about correctness of answers, not whether they literally match expert wording, we focused instead on a binary assessment of whether the generated answer was semantically equivalent to the expert answers. For example, a yes-no answer of ``yes'' versus a generated answer of ``true'' would have \(F_1 = 0\), but we would rate it as correct. For completeness, we also measure and report \(F_1\) scores, but we did not adapt decompositions to maximize \(F_1\).

To keep the task and assessments manageable but meaningful, we sampled 100 questions from \textsc{Qasper} to test both baselines and decompositions that performed substantially better than baselines on the participant flow tasks. We performed light data cleaning before testing, removing special tokens indicating cross-references with natural language equivalents (e.g., converting `BIBREF0' to `[1]') and, as in the original \textsc{Qasper} paper, omitting questions that required tables or figures to answer. 

\subsection{Iterations}

We tested both our select-then-generate baselines and our best-performing decompositions (perplexity selection, with and without few-shot examples from demonstrations) on a sample of \textsc{Qasper} questions. Table \ref{tab:qasper_results} shows the results.

We found that, of the approaches we tested, the best approach on the participant flow questions was also the best approach on \textsc{Qasper}. This approach is using a perplexity classifier followed by a few-shot prompt generated from demonstrations (in this case just other, randomly selected \textsc{Qasper} question-supporting paragraph-answer tuples) and scored 69\% accuracy on our sample. This approach significantly outperformed both the baseline approach of using only the top-1 paragraph from the classifier, then generating the answer (\(38\%\)) and using the Elicit-like perplexity classifier then generating without few-shot demonstrations (\(55\%\)). These findings provide evidence that this decomposition is somewhat robust to domain transfer.

\begin{table}
  \centering
  \begin{tabulars}{X S S }
    \toprule
    \textbf{Method} & \textbf{Accuracy} & $\mathbf{F_1}$ \\ 
    \midrule
    Elicit select-then-generate baseline & \text{38\%} & \text{15\%} \\ 
    + ``not mentioned'' perplexity selection & \text{55\%} & \text{21\%} \\ 
    + ``not mentioned'' perplexity selection + prune & \text{53\%} & \text{20\%} \\ 
    + ``not mentioned'' perplexity selection + auto-few-shot generation & \textbf{69\%} & \textbf{32\%} \\ 
    \bottomrule
  \end{tabulars}
  \caption{Results for the \textsc{Qasper} NLP question-answering task (n=100). The ordering of methods by performance is the same as for the participant flow case study, providing some evidence of generality for decompositions.}
  \label{tab:qasper_results}
\end{table}

\section{Conclusion}

We have described iterated decomposition, a human-in-the-loop workflow for process supervision. We are using this workflow as part of our work on Elicit, the AI Research Assistant. To support this workflow, we built ICE, an open-source debugger for language model programs.

Over time, we expect that similar workflows will be run in a mostly or fully automated fashion, with LMs creating the decompositions, doing error analysis, and adapting the programs. Based on our experience with the case studies, the work needed to get there looks like this:

\begin{enumerate}
  \item Increased iteration speed. Before full automation, we expect to be in the world where human programmers can automate novel complex tasks by improving a task decomposition with many iterations per day.
  \item Complex decompositions. Essentially all decompositions in prior work and to some extent also in our case studies are fairly simple, even when they make hundreds of language model calls.
  \item To accomplish 1 and 2, more sophisticated developer tools that make it faster to go from overall result to the sources of errors, to proposals for how to address them.
  \item Making the dev tools in 3 machine-accessible so that the diagnosis and improvement steps can be learned by LMs.
  \item Reducing the boundary between task design and execution, so that decomposition, debugging, and iteration can happen at runtime as models execute complex tasks.
  \item Distillation and other ways to reduce cost of complex decompositions, so that decompositions are more cost-competitive with end-to-end execution.
\end{enumerate}

Our hope is that we can develop sufficient infrastructure for scalable process supervision so that humans can always in-principle understand how models are solving complex tasks even if eventually almost all decomposition and supervision is done by machines.

\newpage
\renewcommand{\bibsection}{}
\section{References}
\bibliographystyle{unsrtnat}
\bibliography{process-supervision}

\newpage
\section{Appendix}

\subsection{Author contributions}

Justin Reppert and Charlie George ran the participant flow and \textsc{Qasper} case studies.

Ben Rachbach selected and designed the tasks.

Luke Stebbing implemented ICE features.

Maggie Appleton created figures.

Andreas Stuhlmüller ran the placebo case study and wrote the literature review.

Andreas Stuhlmüller, Justin Reppert, and Ben Rachbach wrote the paper.

Jungwon Byun and Andreas Stuhlmüller managed the project.

\subsection{Acknowledgements}

Jason Zukewich helped get ICE off the ground.

Alex Hall started contributing to ICE towards the end of our work on the case studies.

James Brady helped Justin and Luke manage their work.

Paul Christiano and Owain Evans provided helpful thoughts on task decomposition.

Étienne Fortier-Dubois created most of the gold standards used for testing the decompositions.

Beth Barnes, Ethan Perez, Harrison Chase, Jan-Hendrik Kirchner, Jackson Kernion, Jan Leike, Jonathan Uesato, Stanislas Polu, and Owain Evans provided helpful feedback on what to include in the paper.

\subsection{A taxonomy of process supervision}
\label{sec:taxonomy}

In this paper, we use ``process'' to mean the explicit reasoning that the language model performs to reach an answer.

Process can vary along several dimensions, such as:

\begin{itemize}
    \item {\em Causality:} How the reasoning steps are connected and used by the model. For example, chain-of-thought reasoning is a process that model follows on the way to its final answer, but we cannot verify that the reasoning was causally necessary for the final answer. On the other hand, if we first generate reasoning, then feed that output to another call that solves the task without other information that it could use to directly solve the task, we know that the reasoning was part of the causal chain that led to the final answer.
    \item {\em Content:} What the reasoning steps are doing. For example, the model might reason to incorporate more context (such as more sections of a long paper), to use external tools (such as a web search between model calls), or to apply multiple criteria or lines of reasoning to a question.
    \item {\em Designer:} Who decides how to compose reasoning steps. For example, the human developer, the human end user, the language model, or some combination of them (such as the developer providing a set of possible subcompositions and the language model choosing among them).
    \item {\em Purpose:} What the decomposition aims to achieve. For example, the decomposition might aim to improve the performance on the main task, to reflect an ideal or normative reasoning process, or to make the reasoning process more supervisable.
\end{itemize}

In this paper, we focus on decompositions designed by the human developer, with some occasional choices made by the language model. We balance pragmatic decompositions that improve the performance on the main task with decompositions that reflect ideal reasoning and facilitate supervision. Our decompositions mostly improve performance by helping the model use long context more effectively, though some also apply multiple lines of reasoning to subtasks.

Second, ``supervision''. We take supervision to mean checking the outputs or behavior of individual steps in the decomposition.

Supervision can vary along these dimensions:

\begin{itemize}
    \item {\em Goal:} The purpose of supervision could be to: 1) identify and fix failure modes in the decomposition, 2) detect and correct errors during execution, 3) evaluate the final task answer and decide whether to trust it or not, or 4) generate a feedback signal to train the model.
    \item {\em Supervisor:} Who performs the supervision? The developer, the language model, or the user?
    \item {\em Availability of ground truth:} Do we know the correct answer to the task (so we can compare the process outputs to it), or do we not know the correct answer (so we can only judge the process quality by its own logic and consistency)?
\end{itemize}

In this paper, we supervise the process to improve the task decomposition, the supervisor is the developer, and we always have access to the correct answer.

\subsection{Data collection}

For the placebo and participant flow experiments, we evaluated each question against a gold standard answer that we created ourselves. To make these:

\begin{enumerate}
	\item Two workers on the Surge\footnote{\url{https://www.surgehq.ai}} data labeling platform separately wrote what they each believed to be the gold standard answer.
	\item A trusted Ought worker reviewed the two answers from the Surge workers and made the final gold standard answer. If the Surge workers did well, the Ought workers could just accept their answer; more often, the Ought workers had to make substantial changes.
	\item As we used the gold standards for evaluation, we occasionally noticed errors. We fixed these as we caught them.
\end{enumerate}

\subsection{Placebo case study details}

\subsubsection{Task description}
\label{sec:appendix_placebo_task}

We defined a placebo as a substance or intervention, given to the control group, that is meant to mimic the substance or intervention given to the treatment group, but without the ``active ingredient". A placebo attempts to blind participants to whether they were in the treatment or control group.

Besides helping researchers evaluate risk of bias, another reason to pick this task was that it is a fairly easy task. We wanted to try decomposition in a setting where success was likely, and identify any unexpected failure modes.

In the dataset:

\begin{enumerate}
	\item Nearly all papers that used a placebo explicitly said so, so it’s easy to identify these papers (e.g. papers 1 and 3 above)
	\item However, you can’t just look for the word ``placebo", or you’ll get false matches (e.g. paper 4)
	\item Some papers that didn’t use a placebo explicitly said they were open label (paper 2), while others simply described an experimental procedure that doesn’t involve a placebo (paper 5)
	\item Sometimes you can describe the placebo well by just extracting the main description from the paper (paper 1), but sometimes you need to infer from multiple excerpts (paper 3)
\end{enumerate}

We evaluate each trial, rather than the paper. A trial is when participants are randomized into arms that will be tested against each other. Some papers discuss multiple trials, and sometimes some trials are blinded while others are not. There are 61 experiments in our test set across 47 papers.

To evaluate placebo descriptions, we only look at experiments where there is in fact a placebo. If the decompositino said that there is no placebo, we immediately count that description as wrong. Otherwise, we check whether the decomposition's description is reasonably accurate and complete compared to the gold standard.

\subsubsection{Dataset examples}
\label{tab:placebo_examples}

\begin{adjustbox}{max width=\textwidth}
\begin{tabular}{p{1cm}p{12cm}p{4cm}p{4cm}}
\toprule
\textbf{Paper} & 
\textbf{Most relevant quotes} & 
\textbf{Classification} & 
\textbf{Placebo description} \\ 
\midrule
1 & 
"The study was an open randomized controlled clinical trial." & 
No placebo or placebo not mentioned & 
-- \\ 
\addlinespace
2 & 
"Participants were randomized 1:1:1:1 to one of 4 arms: (1) daily TDF beginning at enroLMent, (2) daily placebo beginning at enrollment, (3) daily TDF beginning 9 months after enrollment, and (4) daily placebo beginning 9 months after enrollment (Fig. 1)." \newline
"Participants assigned to TDF were equally or more likely to predict that they were assigned to placebo than to TDF; the opposite was true for placebo participants, suggesting that there was no substantial degree of unblinding." & 
Placebo & 
daily placebo that participants could not distinguish from the treatment \\ 
\addlinespace
3 & 
"Patients in the control condition were put on a waiting list expecting to participate in their peer support group 8 months later." \newline
"The absence of an attention-placebo control condition is a limitation." & 
No placebo or placebo not mentioned & 
-- \\ 
        4 & 
        "The placebo contained the vehicle of the oral azithromycin suspension and was bottled and labeled identically to azithromycin." & 
        Placebo & 
        the vehicle of the oral azithromycin suspension, bottled and labeled identically to azithromycin \\ 
        5 & 
        "Both groups received a preintervention paper survey and a telephone survey 2 to 3 weeks after their clinic visit. The intervention group was offered computer training and received the IP and training summary handout." & 
        \makecell[tl]{No placebo \\ or placebo \\ not mentioned} & 
        -- \\ 
\bottomrule
\end{tabular}
\end{adjustbox}

\subsubsection{Baseline selection failures}
\label{sec:appendix_elicit_selection_failures}

Examples of when selection found mention of a placebo but failed to find the description of the placebo:
    \begin{enumerate}
    \item https://pubmed.ncbi.nlm.nih.gov/27032063/ (Ebner et al. 2016)
        \begin{enumerate}
            \item Selection found only this mention of a placebo: "...That is, while our acute placebo-controlled pharmacological approach provides insight into the effects oxytocin has on resting-state functional connectivity between amygdala and mpfc, our potential for mechanistic explanation is limited..."
            \item Elsewhere the paper describes the placebo: "For the 79 participants whose resting-state functional connectivity data was analyzed, 22 young (50\% female) and 18 older (56\% female) participants were randomly assigned to self-administer via a nasal spray 24 IUs (one puff per nostril) of oxytocin. Eighteen young (50\% female) and 21 older (62\% female) participants self-administered a placebo that contained all ingredients with the exception of the oxytocin at the start of the full study visit."
        \end{enumerate}
    \item https://pubmed.ncbi.nlm.nih.gov/31573637/ (Fowler et al. 2019)
        \begin{enumerate}
            \item Selection found only this mention of a placebo: "...Among the 46 secondary outcomes that were examined in this trial, 43 showed no significant differences between the vitamin c group and the placebo group, although vita-min c compared with placebo was associated with a significant reduction in 28-day all-cause mortality, and with significantly increased icu-free days to day 28 and hospital-free days to day 60..."
            \item Elsewhere the paper describes the placebo: "Patients were randomly assigned to receive intravenous infusion of vitamin C (50 mg/kg in dextrose 5\% in water, n = 84) or placebo (dextrose 5\% in water only, n = 83) every 6 hours for 96 hours."
        \end{enumerate}
    \end{enumerate}

\subsubsection{Final task decomposition}
\label{sec:appendix_placebo_decomposition}

Below is a simplified example of a run of the placebo decomposition. Each step is shown with the answer that it returns. Each child step is used to help return an answer for its parent. \emph{Italic font} describes algorithms, \texttt{monospace font} shows simplified LM prompts. Explore the decomposition in more detail at \url{ought.org/placebo-trace}.

\paragraph{Classification}

\begin{enumerate}
    \item \textbf{Was a placebo used, based on trial arms?} Unclear
\begin{enumerate}
    \item \textbf{What were the trial arms?}: Azithromycin, Placebo
      \begin{enumerate}
        \item \textbf{Rank paragraphs}: Paragraph 12, Paragraph 1,...

\textit{Rank pairs of paragraphs by relevance until we know what the top k paragraphs are}

    \begin{enumerate}
        \item \textbf{Rank pair of paragraphs}: Paragraph 12 > Paragraph 7

\begin{prompt}          
Which of paragraphs 7 and 12 better answers the question, "What were the trial arms (subgroups of participants) of the experiment?"

Paragraph 7: ``Trachoma-control programs have distributed more than 600 million doses of oral azithromycin in an effort to eliminate the ocular strains of chlamydia that cause the disease. 1,2 ..."

Paragraph 12: ``In this cluster-randomized trial, we assigned communities in Malawi, Niger, and Tanzania to four twice-yearly mass distributions of either oral azithromycin (approximately 20 mg per kilogram of body weight) or placebo..."

A: Paragraph 12
\end{prompt}
\end{enumerate}
        \item \textbf{Answer}: Azithromycin, Placebo
\begin{prompt}
Answer the question "What were the trial arms (subgroups of participants) of the experiment?" based on the following paragraphs.

Paragraphs:

In this cluster-randomized trial, we assigned communities in Malawi, Niger, and Tanzania to four twice-yearly mass distributions of either oral azithromycin (approximately 20 mg per kilogram of body weight) or placebo...

A: Azithromycin, Placebo
\end{prompt}
        \end{enumerate}
    \item \textbf{Describe the trial arms}: Communities assigned to receive the oral antibiotic azithromycin, Communities assigned to receive the vehicle of the azithromycin suspension in an identical bottle
    
    \textit{Rank and answer, with relevance determined based on the trial arm we're describing}

        \item \textbf{Do any of the arms look like placebos?}: Yes 

\begin{prompt}
Classify whether the paper used a placebo or not. Err on the side of caution: If you are unsure, answer "Unclear". 

Arm 1: Azithromycin
Description of arm 1: communities assigned to receive the oral antibiotic azithromycin
Arm 2: Placebo
Description of arm 2: communities assigned to receive the vehicle of the azithromycin suspension in an identical bottle

Does the paper use a placebo? Give your reasoning, then answer.

A: Yes
\end{prompt}

    \item \textbf{Could participants tell which arm they were in?}: Yes 

\begin{prompt}
...Followup-question: can participants tell which arm they're in?

A: Yes
\end{prompt}

    \item \textbf{Was a placebo used? (based on arms)}: Unclear
    
\textit{Give the answer from "Naively classify placebo from arms" unless participants can tell which arm they're in, in which case say "Unclear"}

\end{enumerate}
\item \textbf{Was a placebo used, based on the paragraphs?} Yes
\begin{enumerate}
    \item \textbf{Was there a placebo, according to each paragraph?} Yes, Unclear, ...
    \begin{enumerate}
        \item \textbf{Was there a placebo, according to this paragraph?} Yes

\begin{prompt}
Paragraph from paper: The placebo contained the vehicle of the oral azithromycin suspension and was bottled and labeled identically to azithromycin.

Based on the paragraph, did the paper use a placebo? Give your reasoning step by step, then answer.

A: Yes
\end{prompt}
    \end{enumerate}
    \item \textbf{Was a placebo used, aggregate answers from paragraphs?} Yes

\textit{If any paragraph indicated there was no placebo return "No". Then, if any paragraph indicated a placebo, return "Yes". Otherwise, return "No".} 
    \end{enumerate}
    \item \textbf{Was a placebo used, combining answers from arms and paragraphs?} Yes

\textit{In short: say "Yes" if either arms or paragraphs say yes and the other doesn't contradict, else say "No"} 
\end{enumerate}

\paragraph{Description}

For description we again used rank-and-answer. In the example above, this return ``The placebo group received the vehicle of the oral azithromycin suspension. The suspension was bottled and labeled identically to azithromycin.''

\subsubsection{Additional baselines}
\label{sec:appendix_placebo_baselines}
\paragraph{Stuff paper in prompt}

The most obvious language-model baseline is to just stuff as much of the paper as possible in the prompt.

Decomposition outline:

\begin{enumerate}
\item Stuff the prompt with as much of the paper as will fit (a text-davinci-002 prompt plus completion can be 4096 tokens total) and ask the model ``Was this a placebo-controlled study? Let's think step by step:". Then:
\item Classification: Add a classification question after the model’s answer to the original prompt: ``So, to sum up, was this a placebo-controlled study? Answer "Yes", "No", or "Unclear"."
\item Placebo description: Add a description question after the model’s answer to the classification prompt: ``Got it! Please describe the placebo (in one sentence)."
\end{enumerate}

This method does about as well as the Elicit baseline for both classification and description, and seems worse than the task decomposition for both (see Table \ref{tab:placebo_results}; vs.\ the decomposition, p=0.0002 for classification and p=0.06 for description). It's interesting to see that this simple language model approach performs worse than the keyword matching algorithm. Another big weakness of this method is that we don't know what part of the paper the answer is based on, so we can't easily check if it's correct.

\paragraph{Classify by Elicit Prompt “Not Mentioned" Perplexity then Answer with Elicit Prompt}

This approach is discussed in more detail in Section \ref{sec:zero_shot_perplexity}. We include this as an alternative, potentially stronger alternative to the Elicit baseline. In this case, it performs similarly to the Elicit baseline.

\newpage
\subsection{Participant flow case study details}

\subsubsection{CONSORT diagram example}
\label{sec:CONSORT}
    
CONSORT diagrams document the flow of participants through a study. The figure below shows an excerpt of the diagram from \citet{fowler2019}.

\begin{figure}[h]
    \centering
    \includegraphics[width=.5\textwidth]{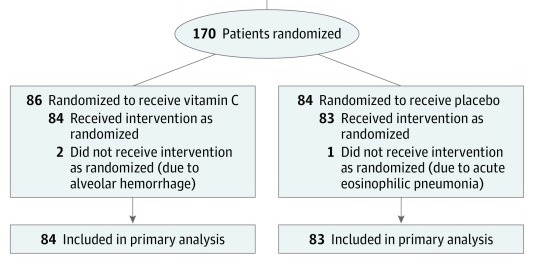}    
\end{figure}

\subsubsection{Dataset examples}
\label{sec:adherence_examples}

The table below shows examples of the most relevant quotes for evaluating adherence in the participant flow task.

\begin{table}[h]
\renewcommand{\arraystretch}{1.3}
\begin{adjustbox}{max width=\textwidth}
\begin{tabularx}{\textwidth}{l X X}
\toprule
\textbf{Paper} & 
\textbf{Most relevant quotes} & 
\textbf{Answer} \\ 
\midrule
1 & 
"All adolescents who were recruited completed all 3 study visits, and               all intervention arm participants initiated gameplay." & 
All participants in the intervention group initiated gameplay. \\
2 & 
"In the Imipramine trial, 49 of the 63 randomized participants completed          6 weeks of treatment (for details, see previous papers 14,15) and 50              of the 63 randomized participants provided ESM data at baseline and               post-intervention and thus were included in the present analyses." & 
49 of the 63 randomized participants across both groups completed             6 weeks of treatment \\ 
\bottomrule
\end{tabularx}
\end{adjustbox}    
\end{table}

\newpage
\subsubsection{Participant flow error analysis}
\label{sec:participant-flow-error-analysis}

\begin{table}[h]
  \centering
  \renewcommand{\arraystretch}{1.3}
  \sisetup{table-format=2.0, table-space-text-post=\%}
  \begin{adjustbox}{max width=\textwidth}
    \begin{tabular}{@{}lSSSS@{}}
      \toprule
      Task: & \multicolumn{2}{c}{\textbf{Adherence}} & \multicolumn{2}{c}{\textbf{Best performance}} \\
      \cmidrule(lr){2-3} \cmidrule(l){4-5}
      Method: & \multicolumn{2}{c}{\textbf{\textit{Baseline}}} & \multicolumn{2}{c}{\textbf{\textit{Best performance}}} \\
      \midrule
      & {n} & {$\%$} & {n} & {$\%$} \\
      False negative & 51 & 80\% & 18 & 45\% \\
      False positive & 6 & 9\% & 12 & 30\% \\
      Hallucination & 4 & 6\% & 0 & 0\% \\
      Missing info & 3 & 5\% & 0 & 0\% \\
      Includes irrelevant info & 1 & 2\% & 6 & 15\% \\
      Answered the wrong question & 0 & 0\% & 4 & 10\% \\
      \midrule
      Total & 64 & 100\% & 40 & 100\% \\
      \bottomrule
    \end{tabular}
  \end{adjustbox}
  \caption{Failure mode counts for adherence}
  \label{tab:adherence_error_analysis}
\end{table}

\begin{table}[h]
    \begin{tabulars}{XXX}
      \toprule
      \textbf{Failure Mode} & 
      \textbf{Decomposition Answer} & 
      \textbf{Gold Answer} \\ 
      \midrule
      False negative & 
      Not mentioned. & 
      52 out of 53 allocated endometrial resections were performed \\ 
      \addlinespace
      False positive & 
      149 children were entered into the study, and each child was followed up for two years. & 
      Not mentioned \\ 
      \addlinespace
      Hallucination & 
      The paper says that the subjects adhered to the treatment regimens. & 
      Across all groups, 1 man dropped out because of compliance problems. \\ 
      \addlinespace
      Missing info & 
      The paper says that both groups had good compliance with all products in each nursing home. & 
      Compliance with oral supplementation was "good", and daily intake averaged about 400 kcal \\ 
      \addlinespace
      Includes irrelevant info & 
      Three of the 43 men dropped out during the treatment phase because of compliance problems (one subject), use of illicit drugs detected on routine drug screen (one subject), and an automobile accident (one subject). & 
      Across all groups, 1 man dropped out because of compliance problems. \\ 
      \addlinespace
      Answered the wrong question & 
      All participants were sober at the start of the experiment, and they were given their drinks and they watched a documentary about South Africa for half an hour to allow the BAC level to reach its maximum. & 
      Adherence was probably 100\%, since the study lasted two hours and was conducted immediately after randomization. \\ 
      \bottomrule
    \end{tabulars}
  \caption{Failure mode examples for adherence}
  \label{tab:adherence_error_analysis_examples}  
\end{table}

\newpage
\subsubsection{Adherence subtask prompt, textbook style}
\label{sec:custom-adherence-prompt-textbook}

We framed some tasks as excerpts from a (nonexistent) textbook on evaluating scientific papers, framing the in-context examples as exercises from this textbook. We found that this discouraged hallucination (since it is unlikely for answers to exercises to go beyond what is available in the supplied text) and allowed us to naturally and didactically specify subtle properties of the task, since textbook examples often build on one another, and it is unlikely for a textbook answer to go beyond what is present in the ``problem''.

\begin{tinyprompt}
From the textbook, "Critically Evaluating Interventional Studies," Chapter 3:

When evaluating the quality of a randomized controlled trial, you should also consider whether any participants dropped out of the study or failed to follow its protocols correctly. This is sometimes called "adherence," "attrition," or "compliance". If too many participants failed to receive the intervention or perform it correctly, for whatever reason, this may damage the internal validity of the study's results.

Unfortunately, papers are often not as clear as they should be when discussing adherence. For simple interventions that are accomplished in one shot (e.g., having a group of college students complete a test in a lab that takes 30 minutes), the study doesn't discuss adherence unless something unusual happened, and we can safely assume that everyone in the sample completed the study. Sometimes studies provide specific numbers or percentages of people who dropped out (attrited), and sometimes they only provide qualitative descriptions, such as saying that adherence was "generally good." Often, papers are genuinely unclear, and we can only conclude that there is not enough information in the paper for us to know anything about adherence or compliance.

Let's look at excerpts from five different papers to see what information, if any, they provide about the study's adherence, attrition, or compliance. We'll have to identify what each extract tells us about adherence (some extracts may only discuss methodology or results, telling us nothing about adherence), and for some, we may have to conclude that the attrition or compliance is simply unclear.

First, consider these three excerpts from a paper studying the Tumaini game:

1. Intervention arm participants completed a 45-minute informational onboarding session, including instructions on the interface, technology, and game content. They were instructed to play at least 1 hour per day for the 16 days of the study and asked not to share their own gameplay profile with others. The game interface allows for 5 additional players' profiles so that others may play without compromising the enrolled player's data. Intervention participants were provided with a phone with the game preloaded and used it at their own pace for the duration of the intervention. Control participants received standard of care, namely no additional intervention beyond any existing sex education from family, school, and peers. No specific data on the content or source of this education were collected from participants. All study smartphones were returned by the participants at the end of the intervention period.

2. Preliminary cleaning of survey data was conducted in MS Excel, with additional cleaning and all analyses completed using SAS version 9.4 (SAS Institute Inc., Cary, NC, USA). All control arm participants were included in analyses. One participant from the intervention arm was removed from analyses of effect at T2 due to delayed completion of the T2 survey. His data were retained for T1-T3 analyses, as he completed all other study activities on time. Descriptive statistics on demographic questions and game feedback questions were computed.

3. We recruited and enrolled 60 adolescent participants. Half of the participants were allocated to the intervention arm. All adolescents who were recruited completed all 3 study visits, and all intervention arm participants initiated gameplay. Participant demographics are presented in Table 3 . There were no significant demographic differences between the two arms. Preliminary calculations of exposure indicate that the intervention arm played Tumaini a mean of approximately 27 hours over the 16 days of the intervention.

Let's think about what each excerpt tells us, if anything, about adherence, attrition or compliance: The first excerpt describes the study's methodology, but does not tell us how many or how well participants followed the instructions, so it does not inform us about adherence. The second excerpt tells us that all control arm participants were included in analysis, but one intervention arm participant was removed from the analysis of effect at T2 but included in the T3 analysis; this is attrition information. The third excerpt says that all participants completed all visits and that all intervention arm participants initiated gameplay; this is adherence information.

Here's all the information in this paper about adherence, attrition, and compliance: All participants completed all visits, and all intervention arm participants initiated gameplay. One intervention arm participant was not included in the T2 analysis but was included in the T3 analysis.

Second, consider these three excerpts from a paper studying the Preschool Situational Self-Regulation Toolkit (PRSIST) Program:

1. All children in their final prior-to-school year in these centers, who attended at least one of the 1-2 assessment days, were invited to participate in this study. There were no further exclusion criteria. Parental consent to participate was provided for 547 3-5-year old children, all of whom were identified as likely to be attending school in the subsequent year. The flow of participants throughout the study is depicted in Figure 1 . At baseline, 473 of these children were assessed (86.5%), with non-participation largely due to absence on the day of assessment. The mean age of this sample was 4.44 years (SD = 0.38, range = 3.20-5.33), with a relative balance of boys and girls (48.2% girls). Children who were identified as of Aboriginal or Torres Strait Islander descent comprised 7.2% of the sample, which is in line with population estimates for this age group (Australian Institute of Health and Welfare (AIHW), 2012). Family income was diverse: 11.9% of families qualified for full childcare benefit subsidies (low income); 65.5% of families qualified for some childcare benefit (low-middle to middle-high income); and 22.7% of families did not qualify for any childcare benefit subsidy (high income). Maternal education levels were also diverse: 9.5% did not complete high school; 9.3% completed only high school; 30.6% had completed a diploma, trade, certificate; 34.6% completed a tertiary degree; and 16.0% a post-graduate qualification. At follow-up, 426 children were assessed, which corresponded to a 90.1% retention rate. Nonparticipation at follow-up was due to the child having left the center or absence on the day of assessment.

2. Based on these patterns of participation, 20 services (80%) were deemed to have met or exceeded the minimum threshold of participation (i.e., completed the professional development modules and met the minimum of three child activities per week). Those that did not participate in the program were a result of: preparations for government assessment and rating (n = 1); substantial illness, maternity leave or turnover of key staff that precluded participation (n = 2); or low-or non-participation for undisclosed reasons (n = 2). Two of these five centers did not participate in any program elements. The other three centers did not engage with professional development modules or induction teleconference call yet completed child activities. Overall, there were good levels of adherence to the program, especially amongst those centers without significant sector-imposed impediments to participation.

3. Inability to conclusively and exclusively provide evidence for one of these possibilities, however, highlights limitations within the current study. That is, although the evaluation was rigorously designed and executed according to CONSORT guidelines, funding considerations limited the roll-out and intervention period to only 6 months. It is possible that a full year of program implementation would yield stronger program effects (see, for example, Schachter, 2015). It is also possible that program effects would be strengthened with stricter adherence to highquality program implementation. While fidelity data indicate good compliance in the frequency and timing of program elements, data are insufficient to evaluate the integrity with which program elements were implemented. While in-person or video fidelity checks were not possible in the current study, this would help monitor adherence. As a researcher-implemented model of delivery would violate our aspiration for a lowcost and barrier-free resource for educators, a plausible middle ground might be a coaching model that supports educators in implementation and adaptation of the program in their context. Lastly, the program was designed with the intention to foster selfregulation in all children, and thus did not focus on instances of dysregulation. However, it is clear that child dysregulation remains a significant concern for educators (Neilsen-Hewett et al., 2019), and future iterations of the program would do well to more explicitly provide support for these children. In guiding such an expansion of the program, there is evidence that children with frequent and severe dysregulation require a different approach to fostering self-regulation, as demonstrated successfully in trauma-informed practice approaches (Holmes et al., 2015). Future studies would also do well to consider implications of differing educator qualifications and experience, whereby different types and levels of support may be needed at varying levels of behavior challenges and educators' skills to address these.

Let's think about what each excerpt tells us, if anything, about adherence, attrition or compliance: The first excerpt includes demographic information about the participants but also reveals that at baseline, 473 of the total sample of 547 children were assessed (with non-participation mostly due to absence), and at follow-up, 426 children were assessed (with non-participation mostly due to the child having left the center or absence), corresponding to a 90.1% retention rate. The second excerpt describes compliance with protocols: 20 of the 25 intervention centers met or exceeded the minimum threshold of participation. The third excerpt describes compliance in the frequency and timing of program elements as "good" but also says that the study did not monitor adherence with in-person or video checks, which would have helped provide a better picture of compliance with the study design.

Here's all the information in this paper about adherence, attrition, and compliance: Of the initial sample of 547 children, 473 were assessed at baseline and 426 at follow-up. While 20 of 25 intervention centers met or exceeded the minimum threshold of participation and the frequency and timing of program elements was good, the study did not monitor adherence with in-person or video checks.

Third, consider these four excerpts from a paper studying Study 2 on depression and psychosis:

1. The intervention was a single session that lasted approximately one hour for participants to provide informed consent, complete a demographic form, watch videos relevant to their study arm, complete the assessments, and be debriefed. Participants in either of the video groups stayed for the full hour, but participants in the control condition who did not watch the video finished in about 50 min. In Study 2, which included two 8 min videos with diagnostic accuracy for both conditions, the protocol required an additional 15 min. Survey data were collected using SurveyCTO (Ver 2.30, Dobility, Inc., Cambridge, MA, USA), an android application, on tablets (www.surveycto.com/accessed on: 19 June 2017). In Study 1, after completion of the video session, participants were invited to participate in the optional qualitative interview to be held within one week.

2. After review of 2nd and 3rd year MBBS student rosters, 18 students were excluded prior to randomization because of being international students not speaking Nepali or having already completed their psychiatry rotation. Among the remaining students, 100 were selected for randomization to one of the three arms. No potential participants refused to participate in this study. An additional six students were excluded at the time of analysis because information on their demographic forms revealed that they were international students whose native language was not Nepali or they had completed their clinical psychiatry rotation; this information had not been up to date in the class rosters at the time of randomization (Figure 1 ). One participant in the service user arm was excluded because of both being an international non-Nepali student and having completed a psychiatry rotation. Demographic characteristics of these participants are in Table 2 . Of note, only three participants indicated that they were primarily interested psychiatry as a specialty (see Figure 2 ). Participants were randomized into one the three conditions: the control group with no video (n = 31, 33%), the didactic video group (n = 31, 33%), and the service user recovery testimonial video group (n = 32; 34%).

3. Due to limited time availability on the part of the researchers and students as well as the exploratory nature of the interviews, only six participants completed interviews. Qualitative results were analyzed from a subset of six students, two women and four men in their third year, who participated in in-depth interviews.

4. For the second study, 248 students were enrolled in first-and second-year MBBS program across the two institutions participating. From roster, 28 students were excluded because of being international or having completed a psychiatry clinical rotation. The remaining 220 students were randomized; however, seven students declined to participate or were unavailable during data collection periods. Therefore, 213 participants were randomly allocated to the following arms: didactic video condition (n = 73), the service user video condition (n = 72), and the no video control condition (n = 75) (Figure 3 ). At the analysis phase, there were additional exclusions because of missing data or identification of exclusion criteria that was not recorded in the school registers. Participant characteristics for each condition are shown in Table 4 .

Let's think about what each excerpt tells us, if anything, about adherence, attrition or compliance. The first excerpt describes the methodology, describing the intervention as taking place in a single one-hour session. This does not tell us anything explicitly about adherence, but it does make it more likely that adherence was high, since participants only had to attend the single session, which is easy to do. The second excerpt says that 18 students were excluded prior to randomization; since this took place before sampling, it is not relevant to adherence. It also says that six students were excluded at the time of analysis because it turned out that they met exclusion criteria. Although this is not adherence strictly speaking, it is important to note when thinking about the validity of the results. The third excerpt says that only six participants completed interviews. The fourth excerpt says that in Study 2, seven students declined to participate or were not available during data collection after randomization of 220 students, and there were additional exclusions at analysis phase because of missing data or identification of exclusion criteria.

Here's all the information in this paper about adherence, attrition, and compliance: This paper does not discuss adherence explicitly. For the video study, six of the 100 randomized students were excluded from analysis, and in the second study, seven of the 220 randomized students declined to participate or were unavailable during data collection periods, with additional students excluded from the analysis because of missing data or identification of exclusion criteria. Only six participants completed interviews.

Fourth, consider these three excerpts from a paper studying antioxidant/anti-inflammatory supplement containing lemon verbena extract and omega-3 fatty acid:

1. Flow chart showing the dropout rate at different timepoints in the study.

2. Forty-eight (48) participants were enrolled for screening evaluation (Fig. 1 ) and after 3 exclusions, 45 participants were randomly assigned either to placebo or nutritional supplement groups, n = 22 and n = 23, respectively. Of these, 14 participants were withdrawn during the study for different reasons; there were 10 dropouts in the placebo group and 4 dropouts in the supplement group (treatment refusal, irregular treatment, starting on medication, or occurrence of an adverse event [AE]). Finally, 31 participants completed the study (12 in the placebo and 19 in the supplement group; Fig. 1 ).

3. Only 1 patient reported an AE (i.e., a heartburn sensation). The subject, who was in the placebo group, stopped the treatment immediately and was excluded from the study (Table 1 ). No major complications were reported by this subject.

Let's think about what each excerpt tells us, if anything, about adherence, attrition or compliance: The first excerpt refers to a flow chart showing the dropout rate, but since we do not have the figure here, we cannot conclude anything from this about the study's attrition. The second excerpt says that there were 10 dropouts in the placebo group of 22 participants and 4 dropouts in the supplement group of 23 participants, meaning that 31 participants out of the initial 45 participants after randomization completed the study. The third excerpt provides more detail for one patient in the placebo group who dropped out, stopping treatment after experiencing a heartburn sensation.

Here's all the the information in this paper about adherence, attrition, and compliance: Ten of the 22 participants in the placebo group dropped out, and 4 of the 23 participants in the supplement group dropped out.

Fifth, consider these {{paragraph_n}} excerpt{{"s" if len(paragraphs) > 1 else ""}} from a paper studying {{intervention}}:

{{paragraphs_to_numbered_list(paragraphs).strip()}}

Let's think about what {{"each" if len(paragraphs) > 1 else "this"}} excerpt tells us, if anything, about adherence, attrition or compliance:
\end{tinyprompt}

\subsubsection{Adherence subtask prompt, Stats Exchange style}
\label{sec:custom-adherence-prompt-stats-exchange}

We experimented with attempting to calculate a final adherence rate where it was available. One possible approach to such calculations is to, e.g., generate Python code then offload the calculations to a Python interpreter. We also found that we could get stronger in-context mathematical reasoning in a zero-shot setting by simulating a StatsExchange post.

\begin{tinyprompt}
A user on Statistics Stack Exchange needs help with this problem:

Use the excerpts from an academic paper to identify how many participants in an intervention completed the intervention or at what rate they completed the intervention. Ignore irrelevant excerpts. If there is not enough information to determine the answer, answer "Unclear".

The intervention is Intracytoplasmic sperm injection (ICSI).

You must decide which excerpts are relevant to your reasoning. Some may be very relevant, and most will be irrelevant.

Excerpts: Thus, -90% of the couples had an embryo transfer and the viable pregnancy rate was 21% for ejaculated, 22% for epididymal and 19% for testicular spermatozoa (with 25-30% multiple pregnancies)....Data forms were sent to all centres that had already taken part in the previous ICSI Task Force survey, as well as to new ICSI centres known from the literature or from national registries....Up to 31 December 1995, 101 centres had submitted their clinical results based on 2 to 2507 cycles (see Table I )....Hence, only 17 of the 101 centres that have submitted ICSI results for 1995 are performing a prospective follow-up of the children and only nine as a part of a special project, while another 46 centres are trying to collect information by contacting the infertility specialist, the paediatrician, or the nurses.

Before giving your answer describe your reasoning in detail. Think about how many people participated in the Intracytoplasmic sperm injection (ICSI). Explain how to deduce an answer the question "What is the adherence rate?" from the relevant excerpts step by step. State your final answer.

The best response from Statistics Stack Exchange was the following:

The quote "Hence, only 17 of the 101 centres that have submitted ICSI results for 1995 are performing a prospective follow-up of the children and only nine as a part of a special project, while another 46 centres are trying to collect information by contacting the infertility specialist, the paediatrician, or the nurses." Talks about the number of centres that performed follow-ups but makes no mention of the number of patients who completed, adhered or complied with the intervention. Therefore it is impossible to tell.

Final answer: Unclear

A user on Statistics Stack Exchange needs help with this problem:

Use the excerpts from an academic paper to identify how many participants in an intervention completed the intervention or at what rate they completed the intervention. Ignore irrelevant excerpts. If there is not enough information to determine the answer, answer "Unclear".

The intervention is malaria chemoprevention (SMC) and long-acting artemisinin-based combination therapies (ACTs).

You must decide which excerpts are relevant to your reasoning. Some may be very relevant, and most will be irrelevant.

Excerpts: A study including approximately 800 children in each group, allowing for a 10% loss to follow-up, would have 90% power to detect a reduction of 20% in the malaria incidence rate between children given a long-acting ACT and a short-acting ACT....A study with 800 children in each group would have greater than 90% power to detect a difference of 25% between children given a long-acting ACT and those given SMC, and very high power to detect larger differences between the SMC group and the short-acting ACT....Caregivers were followed up by CHWs between 4 and 7 days after each SMC administration to ask about adverse events and adherence to the SMC regimen using a standard form....However, it is unlikely that adherence was in reality this high, as during the qualitative component of this study, some caregivers were found to have tablets remaining that had not been administered....Adherence data were collected but implausibly high values of adherence were reported, as seen elsewhere [23].

Before giving your answer describe your reasoning in detail. Think about how many people participated in the malaria chemoprevention (SMC) and long-acting artemisinin-based combination therapies (ACTs). Explain how to deduce an answer the question "What is the adherence rate?" from the relevant excerpts step by step. State your final answer.

The best response from Statistics Stack Exchange was the following:

A quote says "However, it is unlikely that adherence was in reality this high, as during the qualitative component of this study, some caregivers were found to have tablets remaining that had not been administered." This tells us that the follow-up system was not completely accurate, and that some people did not adhere to the SMC regimen. However, we are not given any numbers or percentages, so it is impossible to say what the adherence rate was quantitatively. But this along with "Adherence data were collected but implausibly high values of adherence were reported, as seen elsewhere". Leads us to believe that authors found the adherence to be high but thought is was implausible.

Final answer: The adherence was stated to be high. But authors thought is was implausible.

A user on Statistics Stack Exchange needs help with this problem:

Use the excerpts from an academic paper to identify how many participants in an intervention completed the intervention or at what rate they completed the intervention. Ignore irrelevant excerpts. If there is not enough information to determine the answer, answer "Unclear".

The intervention is pulsed electromagnetic field (PEMF) therapy.

You must decide which excerpts are relevant to your reasoning. Some may be very relevant, and most will be irrelevant.

Excerpts: A total of 286 subjects indicated their interest in the study, and they were sent a health questionnaire with a prepaid return envelope to assess their preliminary eligibility to participate in the study....Twenty-three respondents declined to take part in the trial and 263 respondents returned the questionnaire, 133 of which were excluded from the study due to exclusion criteria....Fourteen patients did not meet the eligibility criteria during the clinical examinations, and eight participants withdrew from the study, thus leaving a sample of 108 participants for the study....Power calculations indicated that a sample of 110 patients (55 in each treatment), assuming a dropout rate of approximately 10%, would provide an 80% (b ¼ 0.20) chance of detecting a 40% (a ¼ 0.05) difference in improvement between the active and sham treatments....Of the 108 patients who started the treatments, nine dropped out during the study....Four patients dropped out during the active treatment period....Five patients dropped out during the sham treatment period....The reasons for those dropouts were amplified overall pain and worsened irritable bowel syndrome (n ¼ 1), lumbar herniated disc (n ¼ 1), cervical herniated disc surgery (n ¼ 1), chronic pneumonia (n ¼ 1), and unwilling to continue (n ¼1; Fig. 1 )....Of the enrolled patients, 92% completed the study....Mean treatment compliance, measured as attendance at all 168 planned, 8-min treatment sessions, was 134 AE 41 (80%) when using the active devices and 131 AE 44 (78%) when using the sham devices....The treatment adherence was high ($80%)....Fourth, considering that high dropout rates are a major issue with crossover design [Mills et al., 2009], in this study the dropout rate was low (8%) despite the relatively long duration of the study (28 weeks).

Before giving your answer describe your reasoning in detail. Think about how many people participated in the pulsed electromagnetic field (PEMF) therapy. Explain how to deduce an answer the question "What is the adherence rate?" from the relevant excerpts step by step. State your final answer.

The best response from Statistics Stack Exchange was the following:

In context of this study the adherence could mean either the completion or compliance rate. We will consider both individually.

For the completion rate we know that of the 108 patient 9 dropout during the course of the study. So 108-9 = 99 complete the study. Hence the completion rate is 99/108 = 91.7%.

For the compliance rate we are told "Mean treatment compliance, measured as attendance at all 168 planned, 8-min treatment sessions, was 134 AE 41 (80%) when using the active devices and 131 AE 44 (78%) when using the sham devices." and that "The treatment adherence was high ($80%)." So the compliance rate is 80% for the active devices and 78% for the sham devices.

Final answer: The completion rate is 91.7%. The compliance rate is 80% for the active devices and 78% for the sham devices.

A user on Statistics Stack Exchange needs help with this problem:

Use the excerpts from an academic paper to identify how many participants in an intervention completed the intervention or at what rate they completed the intervention. Ignore irrelevant excerpts. If there is not enough information to determine the answer, answer "Unclear".

The intervention is {{intervention}}.

You must decide which excerpts are relevant to your reasoning. Some may be very relevant, and most will be irrelevant.

Excerpts: {{"...".join(excerpts)}}

Before giving your answer describe your reasoning in detail. Think about how many people participated in the {intervention}. Explain how to deduce an answer the question "What is the adherence rate?" from the relevant excerpts step by step. State your final answer.

The best response from Statistics Stack Exchange was the following:
\end{tinyprompt}

\newpage
\subsubsection{Decontextualization few-shot prompt}
\label{sec:decontext-prompt-appendix}

\begin{tinyprompt}
Instructions: Enrich each Passage with the Context.

Context: Lisa loves to play practical jokes.

Passage: But sometimes she goes too far.

Rewrite: But sometimes she [Lisa] goes too far.

---

Context: The Super Bowl XLI halftime show took place on February 4, 2007.

Passage: It was headlined by Prince.

Rewrite: It [The Super Bowl XLI halftime show] was headlined by Prince.

---

Context: More than one fifth of the world’s population lives on less than Purchasing Power Parity (PPP) US$1.25 a day, and there is an emerging international consensus that this share should (and can) be driven close to zero by 2030 (1, 2).

Passage: Reaching this objective will require enabling the poorest families, who are often the most marginalized within their villages, to shift from insecure and fragile sources of income to more sustainable income-generating activities.

Rewrite: Reaching this objective [driving the share of the world’s population that lives on less than Purchaing Power Parity (PPP) US$1.25 a day from more than one fifth of the to zero by 2030] will require enabling the poorest families, who are often the most marginalized within their villages, to shift from insecure and fragile sources of income to more sustainable income-generating activities.

---

Context: We present results from randomized control trials (RCTs) in six countries of a particular approach to foster self-employment activities amongst the very poor. Originally designed and implemented by BRAC, a large Bangladeshi NGO that runs several country-wide programs, the ``Graduation" program provides a holistic set of services, including the grant of a productive asset, to the poorest households in a village (referred to by BRAC as the ``ultra-poor"). The beneficiaries [of the Graduation program, the poorest housholds in a village, or the "ultra-poor"] are identified through a participatory process in a village meeting, followed by a verification visit by the organization’s [the implmenter of the "Graduation" program] staff. Selected beneficiaries [among the poorest housholds in a village, or the "ultra-poor"] are then given a productive asset [by the implementer of the "Graduation Program"] that they choose from a list, training and support for the asset they have chosen, as well as general life skills coaching, weekly consumption support for some fixed period, and typically access to savings accounts and health information or services.

Passage: These different activities (plus regular interactions with the households over the course of a year) are designed to complement each other in helping households to start a productive self-employment activity.

Rewrite: These different activities [training and support for the assest they have chosen and received, general life skills coaching, weekly consumption support, and typically access to savings accounts and health information or services] (plus regular interactions with the households over the course of a year) are designed to complement each other in helping households [beneficiaries selected for the "Graduation" program from among the poorest housholds in a village, or the "ultra-poor"] to start a productive self-employment activity.

---

Context: {{context}}

Passage: {{passage}}

Rewrite:
\end{tinyprompt}

\end{document}